\documentclass{article}

\usepackage{microtype}
\usepackage{graphicx}
\usepackage{booktabs} 

\usepackage{hyperref}

\usepackage{bm}
\usepackage{subfiles}
\usepackage{wrapfig}
\usepackage{epsfig}
\usepackage{caption}

\graphicspath{{./images/}}
\DeclareGraphicsExtensions{.pdf,.jpeg,.jpg,.eps,.png}

\newcommand{\INST}[1]{}

\newcommand{\KL}[1]{\textbf{KL}}

\usepackage{amsmath, amsthm, amssymb, xspace}
\usepackage{dsfont}

\usepackage{color}
\definecolor{blue}{rgb}{0,0,.7}
\definecolor{red}{rgb}{.7,0,0}
\definecolor{orange}{rgb}{1,.6,0}
\definecolor{purple}{rgb}{.4,0,.5}
\definecolor{brown}{rgb}{.4,.2,.1}
\definecolor{green}{rgb}{0,.5,0}

\newcommand{\norm}[1]{\left\lVert#1\right\rVert}

\newcommand{\be}{\begin{equation}}
\newcommand{\ee}{\end{equation}}
\newcommand{\bali}{\begin{eqnarray*}}
\newcommand{\eali}{\end{eqnarray*}}
\newcommand{\eq}[1]{\begin{align}#1\end{align}}

\newcommand{\calA}{\mathcal{A}}
\newcommand{\calB}{\mathcal{B}}

\newcommand{\calD}{\mathcal{D}}

\newcommand{\calH}{\mathcal{H}}

\newcommand{\calL}{\mathcal{L}}

\newcommand{\calS}{\mathcal{S}}

\newcommand{\tb}[1]{\textbf{#1}}

\newcommand{\Expect}{\operatorname{\mathbb{E}}}

\newcommand{\bfa}{\textbf{a}}

\newcommand{\bfs}{\textbf{s}}

\newcommand{\bfy}{\textbf{y}}
\newcommand{\bfz}{\textbf{z}}

\DeclareMathOperator*{\argmin}{arg\,min}

\newcommand{\bigbr}[1]{\big(#1\big)}
\newcommand{\Bigbr}[1]{\Big(#1\Big)}

\newcommand{\Metric}{Style-consistency}
\newcommand{\metric}{style-consistency}
\usepackage{subcaption}
\usepackage{cleveref}
\usepackage{sidecap}
\usepackage{tabularx}
\newcolumntype{Y}{>{\centering\arraybackslash}X}

\usepackage[accepted]{icml2020}

\icmltitlerunning{Learning Calibratable Policies using Programmatic Style-Consistency}

\begin{document}

\twocolumn[
\icmltitle{Learning Calibratable Policies using Programmatic Style-Consistency}



\icmlsetsymbol{equal}{*}

\begin{icmlauthorlist}
\icmlauthor{Eric Zhan}{caltech}
\icmlauthor{Albert Tseng}{caltech}
\icmlauthor{Yisong Yue}{caltech}
\icmlauthor{Adith Swaminathan}{microsoft}
\icmlauthor{Matthew Hausknecht}{microsoft}
\end{icmlauthorlist}

\icmlaffiliation{caltech}{California Institute of Technology, Pasadena, CA}
\icmlaffiliation{microsoft}{Microsoft Research, Redmond, WA}

\icmlcorrespondingauthor{Eric Zhan}{ezhan@caltech.edu}

\icmlkeywords{Machine Learning, ICML}

\vskip 0.3in
]



\printAffiliationsAndNotice{}  

\begin{abstract}
We study the problem of controllable generation of long-term sequential behaviors, where the goal is to calibrate to multiple behavior styles simultaneously. In contrast to the well-studied areas of controllable generation of images, text, and speech, there are two questions that pose significant challenges when generating long-term behaviors: how should we specify the factors of variation to control, and how can we ensure that the generated behavior faithfully demonstrates combinatorially many styles? We leverage programmatic labeling functions to specify controllable styles, and derive a formal notion of \metric{} as a learning objective, which can then be solved using conventional policy learning approaches. We evaluate our framework using demonstrations from professional basketball players and agents in the MuJoCo physics environment, and show that existing approaches that do not explicitly enforce style-consistency fail to generate diverse behaviors whereas our learned policies can be calibrated for up to $4^5 (1024)$ distinct style combinations.

\end{abstract}

\section{Introduction}

\begin{figure*}[t]
\centering
    \begin{subfigure}[t]{0.24\linewidth}
        \includegraphics[width=\columnwidth]{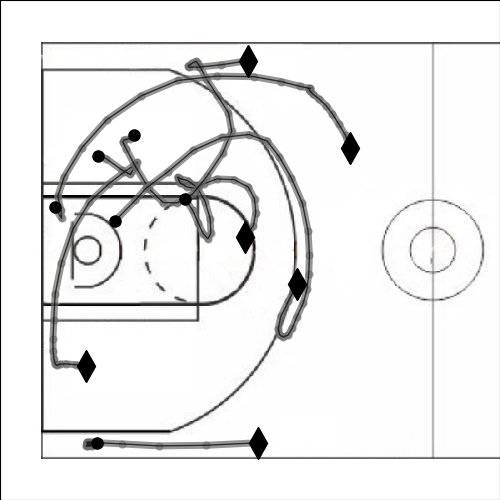}
        \caption{Expert demonstrations}
        \label{fig:bball_intro1}
    \end{subfigure}
    \hfill
    \begin{subfigure}[t]{0.24\linewidth}
        \includegraphics[width=\columnwidth]{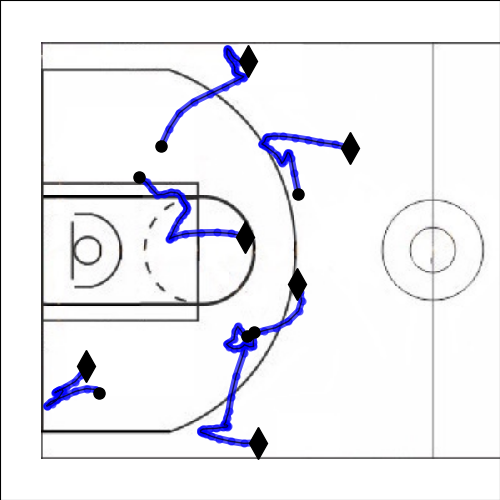}
        \caption{Style: SPEED}
        \label{fig:bball_intro2}
    \end{subfigure}
    \hfill
    \begin{subfigure}[t]{0.24\linewidth}
        \includegraphics[width=\columnwidth]{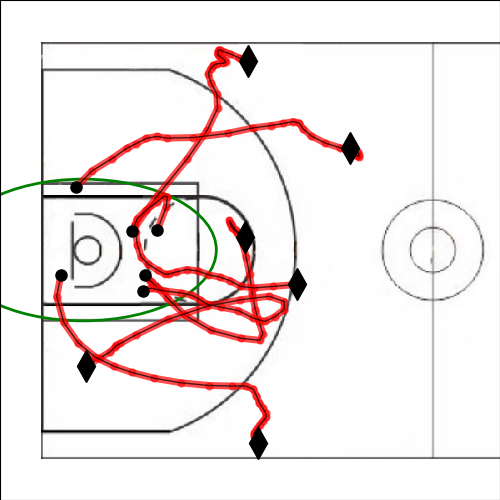}
        \caption{Style: DESTINATION}
        \label{fig:bball_intro3}
    \end{subfigure}
    \hfill
    \begin{subfigure}[t]{0.24\linewidth}
        \includegraphics[width=\columnwidth]{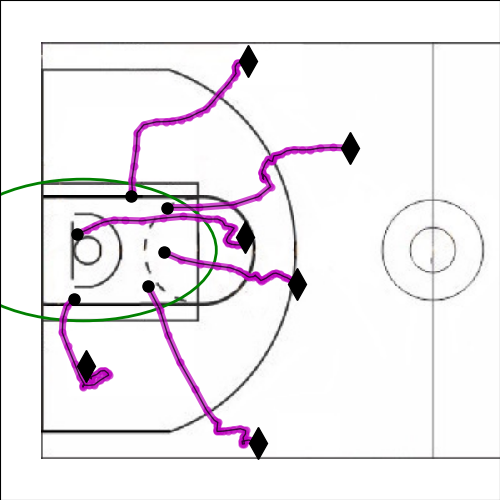}
        \caption{Both styles}
        \label{fig:bball_intro4}
    \end{subfigure}
    \vspace{-5pt}
\caption{Basketball trajectories from policies that are: (a) the expert; (b) calibrated to move at low speeds; (c) calibrated to end near the basket (within green boundary); and (d) calibrated for both (b,c) simultaneously. Diamonds ($\blacklozenge$) and dots ($\bullet$) are initial and final positions.}
\label{fig:bball_intro}
\end{figure*}

\begin{figure*}[t]
\centering
    \begin{subfigure}[t]{0.24\linewidth}
        \includegraphics[width=\columnwidth]{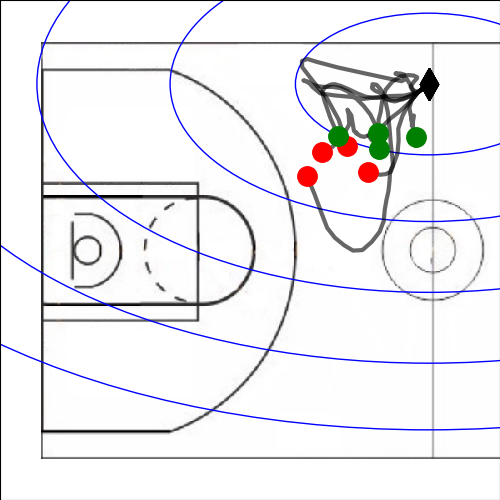}
        \caption{Baseline, low displacement}
        \label{fig:bball_baseline_disp_low}
    \end{subfigure}
    \hfill
    \begin{subfigure}[t]{0.24\linewidth}
        \includegraphics[width=\columnwidth]{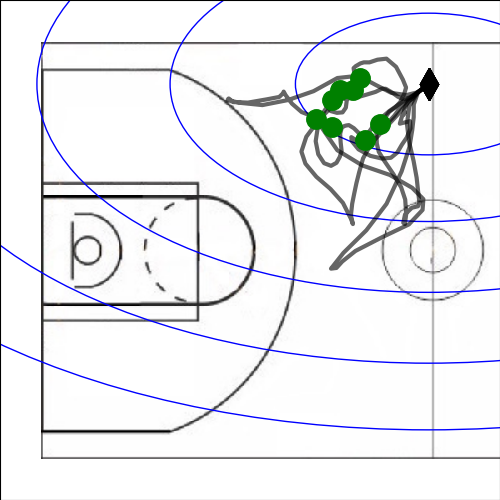}
        \caption{Ours, low displacement}
        \label{fig:bball_style_disp_low}
    \end{subfigure}
    \hfill
    \begin{subfigure}[t]{0.24\linewidth}
        \includegraphics[width=\columnwidth]{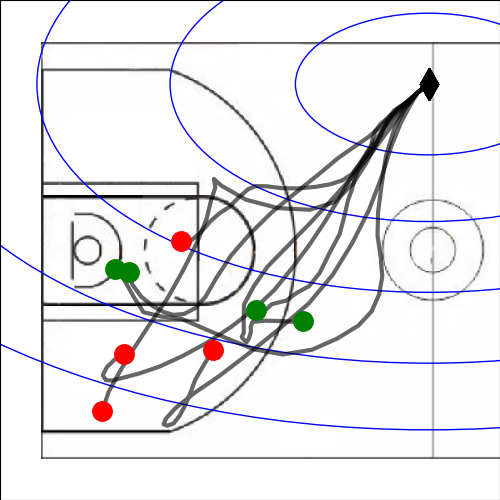}
        \caption{Baseline, high displacement}
        \label{fig:bball_baseline_disp_high}
    \end{subfigure}
    \hfill
    \begin{subfigure}[t]{0.24\linewidth}
        \includegraphics[width=\columnwidth]{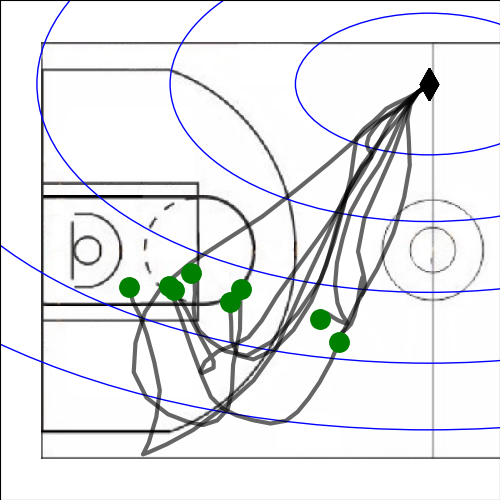}
        \caption{Ours, high displacement}
        \label{fig:bball_style_disp_high}
    \end{subfigure}
    \vspace{-5pt}
\caption{Basketball trajectories sampled from baseline policies and our models calibrated to the style of \texttt{DISPLACEMENT} with 6 classes corresponding to regions separated by blue lines. Diamonds ($\blacklozenge$) and dots ($\bullet$) indicate initial and final positions respectively. Each policy is conditioned on a label class for \texttt{DISPLACEMENT} (low in (a,b), high in (c,d)). Green dots indicate trajectories that are consistent with the style label, while red dots indicate those that are not. Our policy (b,d) is better calibrated for this style than the baselines (a,c).}
\label{fig:bball_ctvae_samples}
\vspace{-10pt}
\end{figure*}

The widespread availability of recorded tracking data   is enabling the study of complex behaviors in many domains, including sports \cite{chen2016learning,le2017coordinated,zhan_19,yeh2019diverse}, video games \cite{atarigrandchallenge,brollcustomizing,hofmann2019minecraft}, laboratory animals \cite{eyjolfsdottir2014detecting,eyjolfsdottir2017learning,branson2009high,johnson2016composing}, facial expressions \cite{suwajanakorn2017synthesizing,taylor2017deep}, commonplace activities such as cooking \cite{nishimura2019frame}, and transportation \cite{bojarski2016end,luo2018fast,li2018diffusion,chang2019argoverse}. A key aspect of modern behavioral datasets is that the behaviors can exhibit very diverse styles (e.g., from multiple demonstrators). For example, Figure \ref{fig:bball_intro1} depicts demonstrations from basketball players with variations in speed, desired destinations, and curvature of movement.

The goal of this paper is to study controllable generation of diverse behaviors by learning to imitate raw demonstrations; or more technically, to develop style-calibrated imitation learning methods. A controllable, or calibratable, policy would enable the generation of behaviors consistent with various styles, such as low movement speed (Figure \ref{fig:bball_intro2}), or approaching the basket (Figure \ref{fig:bball_intro3}), or both styles simultaneously (Figure \ref{fig:bball_intro4}). Style-calibrated imitation learning methods that can yield such policies can be broadly useful to: (a) perform more robust imitation learning from diverse demonstrations \cite{wang_17,brollcustomizing}, (b) enable diverse exploration in reinforcement learning agents \cite{co2018self}, or (c) visualize and extrapolate counterfactual behaviors beyond those seen in the dataset \cite{le2017data}, amongst many other tasks.

Performing style-calibrated imitation is a challenging task. First, what constitutes a ``style''? Second, when can we be certain that a policy is ``calibrated'' when imitating a style? Third, how can we scale policy learning to faithfully generate combinatorially many styles? In related tasks like controllable image generation, common approaches for calibration use adversarial information factorization or mutual information between generated images and user-specified styles (e.g. gender, hair length, etc.) \cite{creswell_18,lample2017fader,chen_16}. However, we find that these \emph{indirect} approaches fall well short of generating calibratable sequential behaviors. Intuitively, the aforementioned objectives provide only indirect proxies for style-calibration. For example, Figure~\ref{fig:bball_ctvae_samples} illustrates that an indirect baseline approach struggles to reliably generate trajectories to reach a certain displacement, even though the dataset contains many examples of such behavior. 

\textbf{Research questions.}
We seek to answer three research questions while tackling this challenge. The first is strategic: since high-level stylistic attributes like movement speed are typically not provided with the raw demonstration data, what systematic form of domain knowledge can we leverage to quickly and cleanly extract highly varied style information from raw behavioral data? The second is formulaic: how can we formalize the learning objective to encourage learning style-calibratable policies that can be controlled to realize many diverse styles?  The third is algorithmic: how do we design practical learning approaches that reliably optimize the learning objective?

\textbf{Our contributions.}
To address these questions, we present a novel framework inspired by \textit{data programming} \citep{ratner_16}, a paradigm in weak supervision that utilizes automated labeling procedures, called labeling functions, to learn without ground-truth labels. In our setting, labeling functions enable domain experts to quickly translate domain knowledge of diverse styles into programmatically generated style annotations. For instance, it is trivial to write programmatic labeling functions for the styles depicted in Figures \ref{fig:bball_intro} \& \ref{fig:bball_ctvae_samples} (speed and destination). Labeling functions also motivate a new learning objective, which we call \textit{programmatic style-consistency}: rollouts generated by a policy calibrated for a particular style should return the same style label when fed to the programmatic labeling function. This notion of style-consistency provides a \emph{direct} approach to measuring how calibrated a policy is, and does not suffer from the weaknesses of indirect approaches such as mutual information estimation. In the basketball example of scoring when near the basket, trajectories that perform correlated events (like turning towards the basket) will not return the desired style label when fed to the labeling function that checks for scoring events. We elaborate on this in Section~\ref{sec:setting}. 

We demonstrate style-calibrated policy learning in Basketball and MuJoCo domains. Our experiments highlight the modularity of our approach -- we can plug-in any policy class and any imitation learning algorithm and reliably optimize for style-consistency using the approach of Section~\ref{sec:approach}. The resulting learned policies can achieve very fine-grained and diverse style-calibration with negligible degradation in imitation quality -- for example, our learned policy is calibrated to $4^5 (1024)$ distinct style combinations in Basketball.

\section{Related Work}
\label{sec:related}

%
Our work combines ideas from policy learning and data programming to develop a weakly supervised approach for more explicit and fine-grained calibration. As such, our work is related to learning disentangled representations and controllable generative modeling, reviewed below.

\textbf{Imitation learning of diverse behaviors}
has focused on unsupervised approaches to infer latent variables/codes that capture behavior styles \citep{li_17, hausman_17, wang_17}. Similar approaches have also been studied for generating text conditioned on attributes such as sentiment or tense \citep{hu17controlled}. A typical strategy  is to maximize the mutual information between the latent codes and trajectories, in contrast to our notion of programmatic style-consistency.

\textbf{Disentangled representation learning}
aims to learn representations where each latent dimension corresponds to exactly one desired factor of variation \citep{bengio_13}. Recent studies \citep{locatello_18} have noted that popular techniques \citep{chen_16, higgins_17, kim_18, chen_18} can be sensitive to hyperparameters and that evaluation metrics can be correlated with certain model classes and datasets, which suggests that fully unsupervised learning approaches may, in general, be unreliable for discovering cleanly calibratable representations. We avoid this roadblock by relying on programmatic labeling functions to provide weak supervision. 

\textbf{Conditional generation}
for images has recently focused on \emph{attribute manipulation} \citep{bao_17, creswell_18, klys_18}, which aims to enforce that changing a label affects only one aspect of the image (similar to disentangled representation learning). We extend these models and compare with our approach in Section~\ref{sec:experiments}. Our experiments suggest that these algorithms do not necessarily scale well into sequential domains.

\textbf{Enforcing consistency in generative modeling}, 
such as cycle-consistency in image generation \citep{zhu2017unpaired}, and self-consistency in hierarchical reinforcement learning \citep{co2018self} has proved beneficial. The former minimizes a discriminative disagreement, whereas the latter minimizes a distributional disagreement between two sets of generated behaviors (e.g., KL-divergence). From this perspective, our style-consistency notion is more similar to the former; however we also enforce consistency over multiple time-steps, which is more similar to the latter.

\textbf{Goal-conditioned policy learning} 
considers policies that take as input the current state along with a desired goal state (e.g., a location), and then must execute a sequence of actions to achieve the goal states. In some cases, the goal states are provided exogenously \cite{zheng2016generating,le2018hierarchical,brollcustomizing,ding2019goal}, and in other cases the goal states are learned as part of a hierarchical policy learning approach \cite{co2018self,sharma2019dynamics} in a way that uses a self-consistency metric similar to our style-consistency approach. Our approach can be viewed as complementary to these approaches as the goal is to study more general notions of consistency (e.g., our styles subsume goals as a special case) as well as to scale to combinatorial joint style spaces. 

\textbf{Hierarchical control via learning latent motor dynamics}
is concerned with recovering a latent representation of motor control dynamics such that one can easily design controllers in the latent space (which then get decoded into actions).  The high level controllers can then be designed afterwards in a pipelined workflow \cite{losey2019controlling,ling2020character,luo2020carl}. The controllers are effective for short time horizons and focus on finding good representations of complex dynamics, whereas we focus on controlling behavior styles that can span longer horizons.

\section{Background: Imitation Learning for Behavior Trajectories}
\label{sec:background}

%
Since our focus is on learning style-calibratable generative policies, for simplicity we develop our approach with the basic imitation learning  paradigm  of behavioral cloning. Interesting future directions include composing our approach with more advanced imitation learning approaches like DAGGER~\citep{DAGGER}, GAIL~\citep{GAIL} as well as with reinforcement learning.

\textbf{Notation.}
Let $\calS$ and $\calA$ denote the environment state and action spaces. At each timestep $t$, an agent observes state $\bfs_t \in \calS$ and executes action $\bfa_t \in \calA$ using a policy $\pi : \calS \rightarrow \calA$. The environment then transitions to the next state $\bfs_{t+1}$ according to a (typically unknown) dynamics function $f : \calS \times \calA \rightarrow \calS$. For the rest of this paper, we  assume $f$ is deterministic; a modification of our approach for stochastic $f$ is included in Appendix \ref{app:stochastic_dynamics}. A trajectory $\tau$ is a sequence of $T$ state-action pairs and the last state: $\tau = \{ (\bfs_t, \bfa_t)\}_{t=1}^T \cup \{ \bfs_{T+1}\}$. Let $\calD$ be a set of $N$  trajectories collected from expert demonstrations. In our experiments, each trajectory in $\calD$ has the same length $T$, but in general this does not need to be the case.

\textbf{Learning objective.}
We begin with the basic imitation learning paradigm of behavioral cloning \citep{syed2008game}.  The goal is to learn a policy that behaves like the pre-collected demonstrations:
\eq{
\pi^* = \argmin_{\pi} \Expect_{\tau \sim \calD} \Big[ \calL^{\text{imitation}}(\tau, \pi) \Big],
\label{eq:imlearn_obj}
}
where $\calL^{\text{imitation}}$ is a loss function that quantifies the mismatch between actions chosen by $\pi$ and those in the demonstrations. Since we are primarily interested in probabilistic or generative policies, we typically use (variants of) negative log-density: $\calL(\tau, \pi) =  \sum_{t=1}^T -\log \pi(\bfa_t|\bfs_t)$, where  $\pi(\bfa_t|\bfs_t)$ is the probability of $\pi$ picking action $\bfa_t$ in $\bfs_t$.

\textbf{Policy class of $\pi$.}
Common model choices for instantiating $\pi$ include sequential generative models like recurrent Neural Networks (RNN) and trajectory variational autoencoders (TVAE). TVAEs introduce a latent variable $\bfz$ (also called a trajectory embedding), an encoder network $q_{\phi}$, a policy decoder $\pi_{\theta}$, and a prior distribution $p$ on $\bfz$. They have been shown to work well in a range of generative policy learning settings \citep{wang_17,ha2018neural,co2018self}, and have the following imitation learning objective:
\begin{align}
\calL^{\text{tvae}}(\tau, \pi_{\theta} ; q_{\phi}) = & \Expect_{q_{\phi}(\bfz|\tau)} \Bigg[ \sum_{t=1}^T -\log \pi_{\theta}(\bfa_t | \bfs_t, \bfz) \Bigg]\nonumber\\ &+ D_{KL} \bigbr{q_{\theta}(\bfz|\tau) || p(\bfz)}.
\label{eq:tvae_obj}
\end{align}
The first term in \eqref{eq:tvae_obj} is the standard \emph{negative log-density} that the policy assigns to trajectories in the dataset, while the second term is the \emph{KL-divergence} between the prior and approximate posterior of trajectory embeddings $\bfz$. The main shortcoming of TVAEs and related approaches, which we address in Sections \ref{sec:setting} \& \ref{sec:approach}, is that the resulting policies cannot be easily calibrated to generate specific styles. For instance, the goal of the trajectory embedding $\bfz$ is to capture all the styles that exist in the expert demonstrations, but there is no guarantee that the embeddings cleanly encode the desired styles in a calibrated way.  Previous work has largely relied on unsupervised learning techniques that either require significant domain knowledge \citep{le2017coordinated}, or have trouble scaling to complex styles commonly found in real-world applications \citep{wang_17,li_17}.

\section{Programmatic \Metric{}}
\label{sec:setting}

%
Building upon the basic setup in Section \ref{sec:background}, 
we focus on the setting where the demonstrations $\calD$ contain diverse behavior styles. To start, let $\bfy \in Y$ denote a single style label (e.g., speed or destination, as shown in Figure \ref{fig:bball_intro}).
Our goal is to learn a policy $\pi$ that can be explicitly calibrated to $\bfy$, i.e., trajectories generated by $\pi(\cdot|\bfy)$ should match the demonstrations in $\calD$ that exhibit style $\bfy$.

Obtaining style labels can be expensive using conventional annotation methods, and unreliable using unsupervised approaches.  We instead  utilize easily programmable labeling functions that automatically produce style labels.  We then formalize a notion of style-consistency as a learning objective, and in Section \ref{sec:approach} describe a practical learning approach.

\textbf{Labeling functions.}
Introduced in the data programming paradigm \citep{ratner_16}, labeling functions programmatically produce weak and noisy labels to learn models on otherwise unlabeled datasets. A significant benefit is that labeling functions are  often simple scripts that can be quickly applied to the dataset, which is much cheaper than manual annotations and more reliable  than unsupervised methods. In our framework, we study behavior styles that can be represented as labeling functions, which we denote $\lambda$, that map trajectories $\tau$  to style labels $\bfy$.
For example:
\eq{
\lambda(\tau) = \mathds{1}\{ \norm{\bfs_{T+1} - \bfs_1}_2 > c \},
\label{eq:lf_example}
}
which distinguishes between trajectories with large (greater than a threshold $c$) versus small total displacement. We experiment with a range of labeling functions, as described in Section \ref{sec:experiments}. Many behavior styles used in previous work can be represented as labeling functions, e.g., agent speed~\citep{wang_17}. 
Multiple labeling functions can be provided at once resulting in a combinatorial space of joint style labels.
We use trajectory-level labels $\lambda(\tau)$ in our experiments, but in general labeling functions can be applied on subsequences $\lambda(\tau_{t:t+h})$ to obtain per-timestep labels, e.g., agent goal~\citep{brollcustomizing}. We can efficiently annotate datasets using labeling functions, which we denote as $\lambda(\calD) = \{ (\tau_i, \lambda(\tau_i)) \}_{i=1}^N$. Our goal can now be phrased as: given $\lambda(\calD)$, train a policy $\pi: \calS \times Y \mapsto \calA$ such that $\pi(\cdot|\bfy)$ is calibrated to styles $\bfy$ found in $\lambda(\calD)$.

\textbf{\Metric{}.}
A key insight in our work is that labeling functions naturally induce a metric for calibration. If a policy $\pi(\cdot|\bfy)$ is calibrated to $\lambda$, we would expect the generated behaviors to be consistent with the label. So, we expect the following loss to be small:
\eq{
 \Expect_{\bfy \sim p(\bfy), \tau \sim \pi(\cdot |\bfy)} \Big[\calL^{\text{style}} \bigbr{\lambda(\tau), \bfy} \Big],
\label{eq:stylecon}
}
where $p(\bfy)$ is a prior over the style labels, and $\tau$ is obtained by executing the style-conditioned policy in the environment. $\calL^{\text{style}}$ is thus a disagreement loss over labels that is minimized at $\lambda(\tau) = \bfy$, e.g., $\calL^{\text{style}} \bigbr{\lambda(\tau), \bfy} = \mathds{1}\{ \lambda(\tau) \neq \bfy \}$ for categorical labels. We refer to \eqref{eq:stylecon} as the \textit{\metric{}} loss, and say that $\pi(\cdot|\bfy)$ is maximally calibrated to $\lambda$ when \eqref{eq:stylecon} is minimized. Our learning objective adds  \eqref{eq:imlearn_obj} with \eqref{eq:stylecon}:
\begin{align}
\pi^* = \argmin_{\pi} & \Expect_{\bigbr{\tau, \lambda(\tau)} \sim \lambda(\calD)} \bigg[ \calL^{\text{imitation}} \Bigbr{\tau, \pi\bigbr{\cdot \mid \lambda(\tau)}} \bigg] \nonumber\\& + \Expect_{\bfy \sim p(\bfy), \tau \sim \pi(\cdot |\bfy)} \Big[\calL^{\text{style}} \bigbr{\lambda(\tau), \bfy} \Big].
\label{eq:full_obj}
\end{align}
\hspace{-0.07in}
The simplest choice for the prior distribution $p(\bfy)$ is the marginal distribution of styles in $\lambda(\calD)$. The first term in \eqref{eq:full_obj} is a standard imitation learning objective and can be tractably estimated using  $\lambda(\calD)$. To enforce \metric{} with the second term, conceptually we need to sample several $\bfy \sim p(\bfy)$, then several rollouts $\tau \sim \pi(\cdot \mid \bfy)$ from the current policy, and query the labeling function for each of them. Furthermore, if $\lambda$ is a non-differentiable function defined over the entire trajectory, as is the case in \eqref{eq:lf_example}, then we cannot simply backpropagate the style-consistency loss. In Section \ref{sec:approach}, we introduce differentiable approximations to more easily optimize the objective in \eqref{eq:full_obj}.

\textbf{Combinatorial joint style space.}
Our notion of \metric{} can be easily extended to optimize for combinatorially-many joint styles when multiple labeling functions are provided. Suppose we have $M$ labeling functions $\{ \lambda_i \}_{i=1}^M$ and corresponding label spaces $\{ Y_i\}_{i=1}^M$. Let $\lambda$ denote $(\lambda_1, \dots, \lambda_M)$ and $\bfy$ denote $(\bfy_1, \dots, \bfy_M)$. \Metric{} loss becomes:
\eq{
\Expect_{\bfy \sim p(\bfy), \tau \sim \pi(\cdot | \bfy)} \Bigg[ \sum_{i=1}^M \calL_i^{\text{style}} \bigbr{\lambda_i(\tau), \bfy_i} \Bigg].
\label{eq:stylecon_multi}
}
Note that \metric{} is optimal when the generated trajectory agrees with \textit{all} labeling functions. Although challenging to achieve, this outcome is most desirable, i.e. $\pi(\cdot|\bfy)$ is calibrated to \textit{all} styles simultaneously.  Indeed, a key metric that we evaluate is how well various learned policies can be calibrated to all styles simultaneously (i.e., loss of 0 only if all styles are calibrated, and loss of 1 otherwise).

\section{Learning Approach}
\label{sec:approach}

%
Optimizing \eqref{eq:full_obj} is challenging due to the long-time horizon and non-differentiability of the labeling functions $\lambda$.\footnote{This issue is not encountered in previous work on style-dependent imitation learning \citep{li_17,hausman_17}, since they use purely unsupervised methods such as maximizing mutual information which is differentiable.} Given unlimited queries to the environment, one could naively employ model-free reinforcement learning, e.g., estimating \eqref{eq:stylecon} using rollouts and optimizing using policy gradient approaches. We instead take a model-based approach, described generically in Algorithm \ref{alg:meta}, that is more computationally-efficient and decomposable (i.e., transparent).  The model-based approach is compatible with batch or offline learning, and we found it particularly useful for diagnosing deficiencies in our algorithmic framework. We first introduce a label approximator for $\lambda$, and then show how to optimize through the environmental dynamics using a differentiable model-based learning approach.

\textbf{Approximating labeling functions.}
To deal with non-differentiability of $\lambda$, we approximate it with a differentiable function $C^\lambda_\psi$ parameterized by $\psi$:
\eq{
\psi^* = \argmin_{\psi} \Expect_{\bigbr{\tau, \lambda(\tau)} \sim \lambda(\calD)} \Big[ \calL^{\text{label}} \bigbr{C^\lambda_{\psi}(\tau), \lambda(\tau)} \Big]
\label{eq:label_approx_obj}
}
Here, $\calL^{\text{label}}$ is a differentiable loss that approximates $\calL^{\text{style}}$, such as cross-entropy loss when $\calL^{\text{style}}$ is the $0/1$ loss. In our experiments we use a RNN to represent $C^\lambda_\psi$. We then modify the \metric{} term in \eqref{eq:full_obj} with $C^\lambda_{\psi^*}$ and optimize:
\begin{align}
\pi^* =  &\argmin_{\pi}  \Expect_{\bigbr{\tau, \lambda(\tau)} \sim \lambda(\calD)} \bigg[ \calL^{\text{imitation}} \Bigbr{\tau, \pi\bigbr{\cdot \mid \lambda(\tau)}} \bigg]\nonumber\\
&\ \ \ \ \ \ \ + \Expect_{\bfy \sim p(\bfy), \tau \sim \pi(\cdot |\bfy)} \Big[ \calL^{\text{label}} \bigbr{C^\lambda_{\psi^*}(\tau), \bfy} \Big].
\label{eq:full_obj_approx}
\end{align}
\vspace{-0.1in}

\begin{algorithm}[tb]
    \caption{Generic recipe for optimizing \eqref{eq:full_obj}}
    \label{alg:meta}
\begin{algorithmic}[1]
    \STATE \textbf{Input}: demonstrations $\calD$, labeling functions $\lambda$
    \STATE construct $\lambda(\calD)$ by applying $\lambda$ on trajectories in $\calD$
    \STATE optimize \eqref{eq:label_approx_obj} to convergence to learn $C^\lambda_{\psi^*}$
    \STATE optimize \eqref{eq:full_obj_approx} to convergence to learn $\pi^*$
\end{algorithmic}
\end{algorithm}

\textbf{Optimizing $\calL^{\text{style}}$ over trajectories. }
The next challenge is to optimize \metric{} over multiple time steps. Consider the labeling function in \eqref{eq:lf_example} that computes the difference between the first and last states. Our label approximator $C^\lambda_{\psi^*}$ may converge to a solution that ignores all inputs except for $\bfs_1$ and $\bfs_{T+1}$. In this case, $C^\lambda_{\psi^*}$ provides no learning signal about intermediate steps. As such, effective optimization of \metric{} in \eqref{eq:full_obj_approx} requires informative learning signals on all actions at every step, which can be viewed as a type of credit assignment problem.

In general, model-free and model-based approaches address this challenge in dramatically different ways and for different problem settings. A model-free solution views this credit assignment challenge as analogous to that faced by reinforcement learning (RL), and repurposes generic reinforcement learning algorithms. Crucially, they assume access to the environment to collect more rollouts under any new policy. A model-based solution does not assume such access and can operate only with the batch of behavior data $\calD$; however they can have an additional failure mode since the learned models may provide an inaccurate signal for proper credit assignment. We choose a model-based approach, while exploiting access to the environment when available to refine the learned models, for two reasons: (a) we found it to be compositionally simpler and easier to debug; and (b) we can use the learned model to obtain hallucinated rollouts of any policy efficiently during training.

\textbf{Modeling dynamics for credit assignment.}
Our model-based approach utilizes a dynamics model $M_{\varphi}$ to approximate the environment's dynamics by predicting the change in state given the current state and action:
\eq{
\varphi^* = \argmin_{\varphi} \Expect_{\tau \sim \calD} \sum_{t=1}^T \calL^{\text{dynamics}} \bigbr{M_{\varphi}(\bfs_t, \bfa_t) , (\bfs_{t+1} - \bfs_t) },
\label{eq:dynamics_obj}
}
where $\calL^{\text{dynamics}}$ is often $L_2$ or squared-$L_2$ loss \citep{nagabandi17neural, luo2019algorithmic}. This allows us to generate trajectories by rolling out: $\bfs_{t+1} = \bfs_t + M_{\varphi}\bigbr{\bfs_t, \pi(\bfs_t)}$. Then optimizing for \metric{} in \eqref{eq:full_obj_approx} would backpropagate through our dynamics model $M_{\varphi}$ and provide informative learning signals to the policy at every timestep.

We outline our model-based approach in Algorithm \ref{alg:model-based-approach}. Lines 12-15 describe an optional step to fine-tune the dynamics model by querying the environment using the current policy (similar to  \citet{luo2019algorithmic}); we found that this  can improve \metric{} in some experiments. In Appendix~\ref{app:stochastic_dynamics} we elaborate how the dynamics model and objective of Eqn~\eqref{eq:dynamics_obj} is changed if the environment is stochastic. 

\textbf{Discussion.}
To summarize, we claim that  style-consistency is an ``objective'' metric to measure the quality of calibration. Our learning approach uses off-the-shelf methods to enforce style-consistency during training. We anticipate several variants of style-consistent policy learning of  Algorithm~\ref{alg:meta} -- for example, using model-free RL, using environment/model rollouts to fine-tune the labeling function approximator, using style-conditioned policy classes, or using other loss functions to encourage imitation quality. Our experiments in Section~\ref{sec:experiments} establish that our style-consistency loss provides a clear learning signal, that no prior approach directly enforces this consistency, and that our approach accomplishes calibration for a combinatorial joint style space.

\begin{algorithm}[t]
    \caption{Model-based approach for Algorithm \ref{alg:meta}}
    \label{alg:model-based-approach}
    \begin{algorithmic}[1]
    \STATE \textbf{Input}: demonstrations $\calD$, labeling function $\lambda$, label approximator $C^\lambda_{\psi}$, dynamics model $M_{\varphi}$
    \STATE $\lambda(\calD) \leftarrow \{ \bigbr{\tau_i, \lambda(\tau_i)} \}_{i=1}^N$ 
    \FOR{$n_{\text{dynamics}}$ iterations}
      \STATE optimize \eqref{eq:dynamics_obj} with batch from $\calD$ 
    \ENDFOR
    \FOR{$n_{\text{label}}$ iterations}
      \STATE optimize \eqref{eq:label_approx_obj} with batch from $\lambda(\calD)$
    \ENDFOR
    \FOR{$n_{\text{policy}}$ iterations}
      \STATE $\calB \leftarrow$ \{ $n_{\text{collect}}$ trajectories using $M_{\varphi}$ and $\pi$ \}
      \STATE optimize \eqref{eq:full_obj_approx} with batch from $\lambda(\calD)$ and $\calB$ 
      \FOR{$n_{\text{env}}$ iterations}
        \STATE $\tau_{\text{env}} \leftarrow$ \{ 1 trajectory using environment and $\pi$ \}
        \STATE optimize \eqref{eq:dynamics_obj} with $\tau_{\text{env}}$ 
      \ENDFOR
    \ENDFOR
\end{algorithmic}
\end{algorithm}

\section{Experiments}
\label{sec:experiments}

%
We first briefly describe our experimental setup and baseline choices, and then discuss our main experimental results.  A full description of experiments is available in Appendix~\ref{app:exp_details}.\footnote{Code is available at: \url{https://github.com/ezhan94/calibratable-style-consistency}.}

\textbf{Data.} 
We validate our framework on two datasets: 1) a collection of professional basketball player trajectories with the goal of learning a policy that generates realistic player-movement, and 2) a Cheetah agent running horizontally in MuJoCo \citep{mujoco} with the goal of learning a policy with calibrated gaits. The former has a known dynamics function: $f(\bfs_t, \bfa_t) = \bfs_t + \bfa_t$, where $\bfs_t$ and $\bfa_t$ are the player's position and velocity on the court respectively; we expect the dynamics model $M_{\varphi}$ to easily recover this function. The latter has an unknown dynamics function (which we learn a model of when approximating style-consistency). We obtain Cheetah demonstrations from a collection of policies trained using \texttt{pytorch-a2c-ppo-acktr} \citep{pytorchrl} to interface with the DeepMind Control Suite's Cheetah domain \citep{dm_control}---see Appendix \ref{app:exp_details} for details.

\textbf{Labeling functions.}
Labeling functions for Basketball include: 1) average \texttt{SPEED} of the player, 2) \texttt{DISPLACEMENT} from initial to final position, 3) distance from final position to a fixed \texttt{DESTINATION} on the court (e.g. the basket), 4) mean \texttt{DIRECTION} of travel, and 5) \texttt{CURVATURE} of the trajectory, which measures the player's propensity to change directions. For Cheetah, we have labeling functions for the agent's 1) \texttt{SPEED}, 2) \texttt{TORSO HEIGHT}, 3) \texttt{BACK-FOOT HEIGHT}, and 4) \texttt{FRONT-FOOT HEIGHT} that can be trivially computed from trajectories extracted from the environment.

We threshold the aforementioned labeling functions into categorical labels (leaving real-valued labels for future work)
and use \eqref{eq:stylecon} for \metric{} with $\calL^{\text{style}}$ as the $0/1$ loss. We use cross-entropy for $\calL^{\text{label}}$ and list all other hyperparameters in Appendix \ref{app:exp_details}.

\textbf{Metrics.} We will primarily study two properties of the learned models in our experiments -- imitation quality, and style-calibration quality. For measuring imitation quality of generative models, we report the \emph{negative log-density} term in \eqref{eq:tvae_obj}, also known as the reconstruction loss term in VAE literature \citep{kingma2014auto, ha2018neural}, which corresponds to how well the policy can reconstruct trajectories from the dataset. 

To measure style-calibration, we report \metric{} results as $1 - \calL^{\text{style}}$ in \eqref{eq:stylecon} so that all results are easily interpreted as accuracies. In Section~\ref{sec:exp_noisy}, we find that \metric{} indeed captures a reasonable notion of calibration -- when the labeling function is inherently noisy and style-calibration is hard, \metric{} correspondingly decreases. In Section~\ref{sec:exp_tradeoff}, we find that the goals of imitation (as measured by negative log-density) and calibration (as measured by \metric{}) may not always be aligned -- investigating this trade-off is an avenue for future work.

\textbf{Baselines.} 
Our main experiments use TVAEs as the underlying policy class.
In Section~\ref{sec:exp_rnn}, we also experiment with an RNN policy class. We compare our approach, CTVAE-style, with 3 baselines:
\vspace{-0.1in}
\begin{enumerate}
    \item \textbf{CTVAE}: conditional TVAEs \citep{wang_17}.
    \vspace{-0.05in}
    \item \textbf{CTVAE-info}: CTVAE with information factorization \citep{creswell_18}, \emph{indirectly} maximizes \metric{} by removing correlation of $\bfy$ with $\bfz$.
    \vspace{-0.05in}
    \item \textbf{CTVAE-mi}: CTVAE with mutual information maximization between style labels and trajectories. This is a supervised variant of unsupervised models \citep{chen_16, li_17}, and also requires learning a dynamics model for sampling policy rollouts.
\end{enumerate}

Detailed descriptions of baselines are in Appendix \ref{app:baselines}. All baseline models build upon TVAEs, which are also conditioned on a latent variable (see Section \ref{sec:background}) and only fundamentally differ in how they encourage the calibration of policies to different style labels. We highlight that the underlying model choice is orthogonal to our contributions; our framework is compatible with other policy models (see Section \ref{sec:exp_rnn}).

\textbf{Model details.}
We model all trajectory embeddings $\bfz$ as a diagonal Gaussian with a standard normal prior. Encoder $q_{\phi}$ and label approximators $C_{\psi}^{\lambda}$ are bi-directional GRUs \citep{gru} followed by linear layers. Policy $\pi_{\theta}$ is recurrent for basketball, but non-recurrent for Cheetah. The Gaussian log sigma returned by $\pi_{\theta}$ is state-dependent for basketball, but state-independent for Cheetah. For Cheetah, we made these choices based on prior work in MuJoCo for training gait policies \citep{wang_17}. For Basketball, we observed a lot more variation in the 500k demonstrations so we experimented with a more flexible model. See Appendix \ref{app:exp_details} for hyperparameters.

\subsection{How well can we calibrate policies for single styles?}
\label{sec:exp1}

%
We first threshold labeling functions into 3 classes for Basketball and 2 classes for Cheetah; the marginal distribution $p(\bfy)$ of styles in $\lambda(\calD)$ is roughly uniform over these classes. Then we learn a policy $\pi^*$ calibrated to each of these styles. Finally, we generate rollouts from each of the learned policies to measure \metric{}. Table \ref{tab:exp_1} compares the median \metric{} (over 5 seeds) of learned policies. For Basketball, CTVAE-style significantly outperforms baselines and achieves almost perfect \metric{} for 4 of the 5 styles. For Cheetah, CTVAE-style outperforms all baselines, but the absolute performance is lower than for Basketball -- we conjecture that this is due to the complex environment dynamics that can be challenging for model-based approaches. Figure \ref{fig:bball_lf_dest} in Appendix \ref{app:exp_results} shows a visualization of our CTVAE-style policy calibrated for \texttt{DESTINATION(net)}.

We also consider cases in which labeling functions can have several classes and non-uniform distributions (i.e. some styles are more/less common than others). We threshold \texttt{DISPLACEMENT} into 6 classes for Basketball and \texttt{SPEED} into 4 classes for Cheetah and compare the policies in Table \ref{tab:exp_2}. In general, we observe degradation in overall \metric{} accuracies as the number of classes increase. However, CTVAE-style policies still consistently achieve better \metric{} than baselines in this setting.

We visualize and compare policies calibrated for 6 classes of \texttt{DISPLACEMENT} in Figure \ref{fig:bball_ctvae_samples}. In Figure \ref{fig:bball_style_disp_low} and \ref{fig:bball_style_disp_high}, we see that our CTVAE-policy  (0.92 \metric{}) is effectively calibrated for styles of low and high displacement, as all trajectories end in the correct corresponding regions (marked by the green dots). On the other hand, trajectories from a baseline CTVAE model (0.70 \metric{}) in Figure \ref{fig:bball_baseline_disp_low} and \ref{fig:bball_baseline_disp_high} can sometimes end in the wrong region corresponding to a different style label (marked by red dots). These results suggest that incorporating programmatic \metric{} while training via \eqref{eq:full_obj_approx} can yield good qualitative and quantitative calibration results.

\begin{table}
  \centering
  \begin{subtable}[t]{0.48\textwidth}
     \small
     \centering
     \begin{tabularx}{\linewidth}{ |l| *{5}{Y|} }
        \hline
        \tb{Model} & \tb{Speed} & \tb{Disp.} & \tb{Dest.} & \tb{Dir.} & \tb{Curve} \\
        \hline
        CTVAE      & 83 & 72 & 82 & 77 & 61 \\
        CTVAE-info & 84 & 71 & 79 & 72 & 60 \\
        CTVAE-mi   & 86 & 74 & 82 & 77 & 72 \\
        \hline
        CTVAE-style & \tb{95} & \tb{96} & \tb{97} & \tb{97} & \tb{81} \\
        \hline
    \end{tabularx}
    \caption{\Metric{} for labeling functions in Basketball.}
    \label{tab:exp_1_bball}
  \end{subtable}
  
  \begin{subtable}[t]{0.48\textwidth}
     \small
     \centering
     \vspace{0.02in}
     \begin{tabularx}{0.85\linewidth}{ |l| *{4}{Y|} }
        \hline
        \tb{Model} & \tb{Speed} & \tb{Torso} & \tb{BFoot} & \tb{FFoot} \\
        \hline
        CTVAE      & 59 & 63 & 68 & 68 \\
        CTVAE-info & 57 & 63 & 65 & 66 \\
        CTVAE-mi   & 60 & 65 & 65 & 70 \\
        \hline
        CTVAE-style & \tb{79} & \tb{80} & \tb{80} & \tb{77} \\
        \hline
    \end{tabularx}
     \caption{\Metric{} for labeling functions in Cheetah.}
     \label{tab:exp_1_cheetah}
  \end{subtable}
  \caption{\textbf{Individual Style Calibration:} \Metric{} ($\times 10^{-2}$, median over 5 seeds) of policies evaluated with 4,000 Basketball and 500 Cheetah rollouts. Trained separately for each style, CTVAE-style policies outperform baselines for all styles in Basketball and Cheetah environments.}
  \label{tab:exp_1}
\end{table}

\begin{table}[t]
\small
\centering
    \begin{tabularx}{\columnwidth}{ |l| *{6}{Y|} }
        \cline{2-7}
        \multicolumn{1}{c|}{} 
        & \multicolumn{4}{c|}{\tb{Basketball}}
        & \multicolumn{2}{c|}{\tb{Cheetah}} \\
        \hline
        \tb{Model} & \tb{2 class} & \tb{3 class} & \tb{4 class} & \tb{6 class} & \tb{3 class} & \tb{4 class}  \\
        \hline
        CTVAE      & 92 & 83 & 79 & 70 & 45 & 37 \\
        CTVAE-info & 90 & 83 & 78 & 70 & 49 & 39 \\
        CTVAE-mi   & 92 & 84 & 77 & 70 & 48 & 37 \\
        \hline
        CTVAE-style & \tb{99} & \tb{98} & \tb{96} & \tb{92} & \tb{59} & \tb{51} \\
        \hline
    \end{tabularx}
    \vspace{-0.1in}
\caption{\textbf{Fine-grained Style-consistency:} ($\times 10^{-2}$, median over 5 seeds) Training on labeling functions with more classes (\texttt{DISPLACEMENT} for Basketball, \texttt{SPEED} for Cheetah) yields increasingly fine-grained calibration of behavior. Although CTVAE-style degrades as the number of classes increases, it outperforms baselines for all styles.}
\label{tab:exp_2}
\end{table}

\begin{table}[t]
    \centering
    \begin{subtable}[t]{0.48\textwidth}
     \small
     \centering
     \setlength{\tabcolsep}{1pt}
     \begin{tabularx}{\columnwidth}{ |l| *{5}{Y|} }
        \hline
        & \tb{2 style} & \tb{3 style} & \tb{4 style} & \tb{5 style} & \tb{5 style}  \\
        & \tb{3 class} & \tb{3 class} & \tb{3 class} & \tb{3 class} & \tb{4 class}  \\
        \tb{Model} & (8) & (27) & (81) & (243) & (1024)  \\
        \hline
        CTVAE      & 71 & 58 & 50 & 37 & 21 \\
        CTVAE-info & 69 & 58 & 51 & 32 & 21 \\
        CTVAE-mi   & 72 & 56 & 51 & 30 & 21 \\
        \hline
        CTVAE-style & \tb{93} & \tb{88} & \tb{88} & \tb{75} & \tb{55} \\
        \hline
    \end{tabularx}
    \caption{\Metric{} for labeling functions in Basketball.}
    \label{tab:exp_3_bball}
    \end{subtable}
    
    \begin{subtable}[t]{0.48\textwidth}
     \small
     \centering
     \vspace{0.02in}
     \begin{tabularx}{0.58\columnwidth}{ |l| *{6}{Y|} }
        \hline
        & \tb{2 style} & \tb{3 style}  \\
        & \tb{2 class} & \tb{2 class}  \\
        \tb{Model} & (4)  & (8)  \\
        \hline
        CTVAE      & 41 & 28 \\
        CTVAE-info & 41 & 27 \\
        CTVAE-mi   & 40 & 28 \\
        \hline
        CTVAE-style & \tb{54} & \tb{40} \\
        \hline
     \end{tabularx}
     \caption{\Metric{} for labeling functions in Cheetah.}
     \label{tab:exp_3_cheetah}
    \end{subtable}
\caption{\textbf{Combinatorial Style-consistency:} ($\times 10^{-2}$, median over 5 seeds) Simultaneously calibrated to joint styles from multiple labeling functions, CTVAE-style policies significantly outperform all baselines. The number of distinct style combinations are in brackets. The most challenging experiment for basketball calibrates for $1024$ joint styles (5 labeling functions, 4 classes each), in which CTVAE-style has a $+161\%$ improvement in \metric{} over the best baseline.}
\label{tab:exp_3}
\vspace{-10pt}
\end{table}

\subsection{Can we calibrate for combinatorial joint style spaces?}
\label{sec:exp_multi}

\begin{figure*}
\vspace{-0.05in}
  \begin{minipage}[c]{0.75\textwidth}
    \begin{center}
        \begin{figure}[H]
        \begin{center}
        \begin{subfigure}[t]{0.325\columnwidth}
        \centering
        \includegraphics[width=.8\columnwidth]{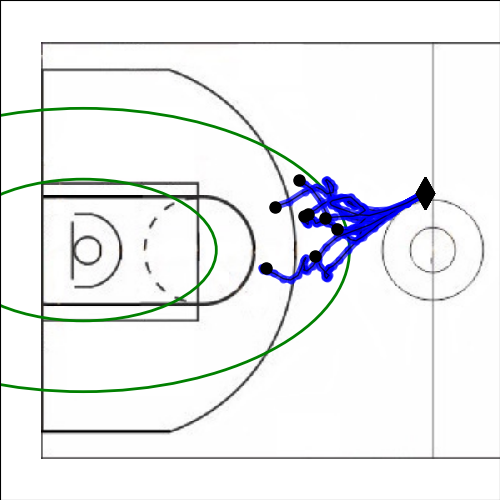}
        \vspace{-0.05in}
        \caption{Label class 0 (slow)}
        \label{fig:bball_2lf_0}
        \end{subfigure}
        \begin{subfigure}[t]{0.325\columnwidth}
        \centering
        \includegraphics[width=.8\columnwidth]{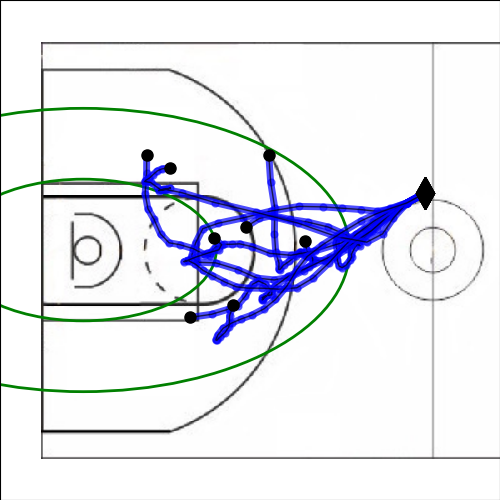}
        \vspace{-0.05in}
        \caption{Label class 1 (mid)}
        \label{fig:bball_2lf_1}
        \end{subfigure}
        \begin{subfigure}[t]{0.325\columnwidth}
        \centering
        \includegraphics[width=.8\columnwidth]{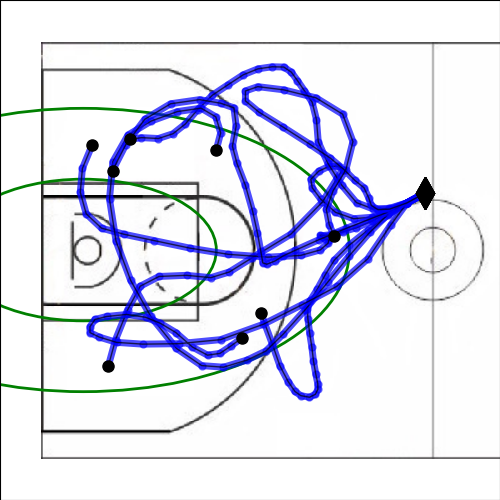}
        \vspace{-0.05in}
        \caption{Label class 2 (fast)}
        \label{fig:bball_2lf_2}
        \end{subfigure}
        \end{center}
        \end{figure}
    \end{center}
  \end{minipage}\hfill
  \begin{minipage}[c]{0.24\textwidth}
    \vspace{10pt}
    \caption{CTVAE-style rollouts calibrated for 2 styles: label class 1 of \texttt{DESTINATION(net)} (see Figure \ref{fig:bball_lf_dest} in Appendix \ref{app:exp_results}) and each class for \texttt{SPEED}, with 0.93 \metric{}. Diamonds ($\blacklozenge$) and dots ($\bullet$) indicate initial and final positions.}
    \label{fig:bball_2lf}
  \end{minipage}
  \vspace{-0.05in}
\end{figure*}

%
We now consider combinatorial \metric{} as in \eqref{eq:stylecon_multi}, which measures the \metric{} with respect to \textit{all} labeling functions simultaneously. For instance, in Figure \ref{fig:bball_2lf}, we calibrate to both  terminating close to the net and also the speed at which the agent moves towards the target destination; if either style is not calibrated then the joint style is not calibrated.  In our experiments, we evaluated up to 1024 joint styles.

Table \ref{tab:exp_3} compares the \metric{} of policies simultaneously calibrated for up to 5 labeling functions for Basketball and 3 labeling functions for Cheetah. This is a very difficult task, and we see that \metric{} for baselines degrades significantly as the number of joint styles grows combinatorially. On the other hand, our CTVAE-style approach experiences only a modest decrease in \metric{} and is still significantly better calibrated (0.55 \metric{} vs. 0.21 best baseline \metric{} in the most challenging experiment for Basketball). We visualize a CTVAE-style policy calibrated for two styles in Basketball with \metric{} 0.93 in Figure \ref{fig:bball_2lf}. CTVAE-style outperforms baselines in Cheetah as well, but there is still room for improvement to optimize \metric{} better in future work.

\subsection{What is the trade-off between \metric{} and imitation quality?}
\label{sec:exp_tradeoff}

%
In Table \ref{tab:tradeoff}, we investigate whether CTVAE-style's superior \metric{} comes at a significant cost to imitation quality, since we optimize both in \eqref{eq:full_obj}. For Basketball, high \metric{} is achieved without any degradation in imitation quality. For Cheetah, negative log-density is slightly worse; a followup experiment in Table \ref{tab:cheetah_tradeoff} in Appendix \ref{app:exp_results} shows that we can improve imitation quality with more training, sometimes with modest decrease to \metric{}.

\subsection{Is our framework compatible with other policy classes for imitation?}
\label{sec:exp_rnn}

%
We highlight that our framework introduced in Section \ref{sec:approach} is compatible with any policy class. In this experiment, we optimize for \metric{} using a simpler model for the policy and show that \metric{} is still improved. In particular, we use an RNN and calibrate for \texttt{DESTINATION} in basketball. In Table \ref{tab:exp_rnn_main}, we see that \metric{} is improved for the RNN model without any significant decrease in imitation quality.

\begin{table}[t]
    \centering
    \small
    \begin{tabularx}{0.8\linewidth}{ |l| *{4}{Y|} }
        \cline{2-5}
        \multicolumn{1}{c|}{} 
        & \multicolumn{2}{c|}{\tb{Basketball}}
        & \multicolumn{2}{c|}{\tb{Cheetah}} \\
        \hline
        \tb{Model} & $D_{KL}$ & NLD & $D_{KL}$ & NLD \\
        \hline
        TVAE        & 2.55 & -7.91 & 29.4 & -0.60 \\
        CTVAE       & 2.51 & -7.94 & 29.3 & -0.59 \\
        CTVAE-info  & 2.25 & -7.91 & 29.1 & -0.58 \\
        CTVAE-mi    & 2.56 & -7.94 & 28.5 & -0.57 \\
        \hline
        CTVAE-style & 2.27 & -7.83 & 30.1 & -0.28 \\
        \hline
    \end{tabularx}
    \vspace{-0.1in}
    \caption{KL-divergence and negative log-density per timestep for TVAE models (lower is better). CTVAE-style is comparable to baselines for Basketball, but is slightly worse for Cheetah.}
    \label{tab:tradeoff}
    \vspace{-0.05in}
\end{table}

\begin{table}[t]
\centering
    \small
    \begin{tabularx}{\columnwidth}{ |l| *{4}{Y|} }
        \cline{2-4}
        \multicolumn{1}{c|}{} 
        & \multicolumn{3}{c|}{\tb{\Metric{} $\uparrow$}}
        & \multicolumn{1}{c}{} \\
        \hline
        \tb{Model} & \tb{Min} & \tb{Median} & \tb{Max} & NLD $\downarrow$ \\
        \hline
        RNN        & 79 & 80 & 81 & \tb{-7.7} \\
        RNN-style  & \tb{81} & \tb{91} & \tb{98} & -7.6 \\
        \hline
    \end{tabularx}
    \vspace{-0.1in}
\caption{\Metric{} of RNN policy model ($10^{-2}$, 5 seeds) for \texttt{DESTINATION} in basketball. Our approach improves \metric{} without significantly decreasing imitation quality.}
\label{tab:exp_rnn_main}
\vspace{-10pt}
\end{table}

\subsection{What if labeling functions are noisy?}
\label{sec:exp_noisy}

%
So far, we have demonstrated that our method optimizing for \metric{} directly can learn policies that are much better calibrated to styles, without a significant degradation in imitation quality. However, we note that the labeling functions used thus far are assumed to be perfect, in  that they capture exactly the style that we wish to calibrate. In practice, domain experts may specify labeling functions that are noisy; we simulate that scenario in this experiment.

In particular, we create noisy versions of labeling functions in Table \ref{tab:exp_1} by adding Gaussian noise to computed values before applying the  thresholds. The noise will result in some label disagreement between noisy and true labeling functions (Table \ref{tab:exp_noise_mismatch} in Appendix \ref{app:exp_results}). This resembles the scenario in practice where domain experts can mislabel a trajectory, or have disagreements. We train CTVAE-style models with noisy labeling functions and compute \metric{} using the true labeling functions without noise. Intuitively, we expect the relative decrease in \metric{} to scale linearly with the label disagreement.

\begin{figure}
    \centering
    \includegraphics[width=0.99\columnwidth]{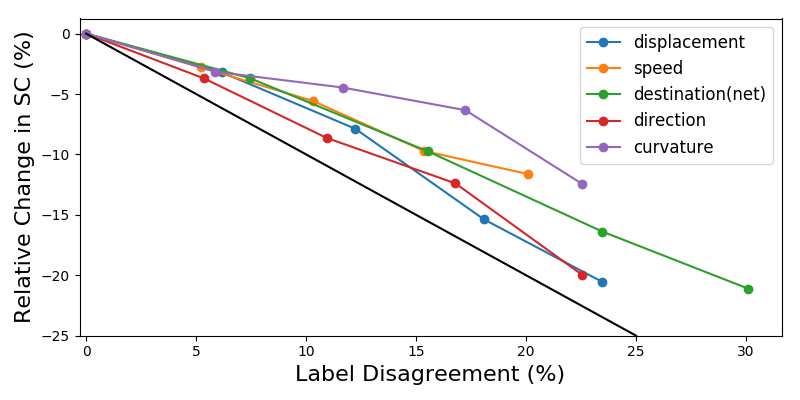}
    \vspace{-0.1in}
    \caption{Relative change of \metric{} for CTVAE-style policies trained with noisy labeling functions, which are created by injecting noise with mean 0 and standard deviation $c \cdot \sigma$ for $c \in \{ 1, 2, 3, 4 \}$ before applying thresholds to obtain label classes. The x-axis is the label disagreement between noisy and true labeling functions. The y-axis is the median change (5 seeds) in \metric{} using the true labeling functions without noise, relative to Table \ref{tab:exp_1}. The relationship is generally linear and better than a one-to-one dependency (i.e. if $X\%$ label disagreement leads to $-X\%$ relative change, indicated by the black line). See Table \ref{tab:exp_noise_mismatch} and \ref{tab:exp_noise} in the Appendix \ref{app:exp_results} for more details.}
    \label{fig:exp_noise}
\end{figure}

Figure \ref{fig:exp_noise} shows that the median relative decrease in \metric{} of our CTVAE-models scales linearly with label disagreement. Our method is also somewhat robust to noise, as $X\%$ label disagreement results in better than $X\%$ relative decrease in \metric{} (black line in Figure \ref{fig:exp_noise}). Directions for  future work include combining multiple noisy labeling functions together to improve \metric{} with respect to a ``true'' labeling function.

\section{Conclusion and Future Work}
\label{sec:conclusion}

%
We propose a novel framework for imitating diverse behavior styles while also  calibrating to desired styles. Our framework leverages labeling functions to tractably represent styles and introduces programmatic \metric{}, a metric that allows for fair comparison between calibrated policies. Our experiments demonstrate strong empirical calibration results.

We believe that our framework lays the foundation for many directions of future research. First, can one model more complex styles not easily captured with a single labeling function (e.g. aggressive vs. passive play in sports) by composing simpler labeling functions (e.g. max speed, distance to closest opponent, number of fouls committed, etc.), similar to \citep{ratner_16, bach_17}? Second, can we use these per-timestep labels to model transient styles, or simplify the credit assignment problem when learning to calibrate? Third, can we blend our programmatic supervision with unsupervised learning approaches to arrive at effective semi-supervised solutions? Fourth, can we use model-free approaches to further optimize self-consistency, e.g., to fine-tune from our model-based approach? Finally, can we integrate our framework with reinforcement learning to also optimize for environmental rewards?

\section*{Acknowledgements}

This research is supported in part by NSF \#1564330, NSF \#1918655, DARPA PAI, and gifts from Intel, Activision/Blizzard and Northrop Grumman. Basketball dataset was provided by STATS.

\begin{small}
\bibliography{references}
\bibliographystyle{icml2020}
\end{small}

\appendix
\appendix

\onecolumn

\section{Baseline Policy Models}
\label{app:baselines}

%
\paragraph{1) Conditional-TVAE (CTVAE).}
The conditional version of TVAEs optimizes:
\eq{
\calL^{\text{ctvae}}(\tau, \pi_{\theta}; q_{\phi}) = \Expect_{q_{\phi}(\bfz|\tau. \bfy)} \Bigg[ \sum_{t=1}^T -\log \pi_{\theta}(\bfa_t | \bfs_t, \bfz, \bfy) \Bigg] + D_{KL} \bigbr{q_{\theta}(\bfz|\tau, \bfy) || p(\bfz)}.
\label{eq:ctvae_obj}
}

\paragraph{2) CTVAE with information factorization (CTVAE-info).}
\citep{creswell_18, klys_18} augment conditional-VAE models with an auxiliary network $A_{\psi}(\bfz)$ which is trained to predict the label $\bfy$ from $\bfz$, while the encoder $q_{\phi}$ is also trained to minimize the accuracy of $A_{\psi}$. This model \textit{implicitly} maximizes self-consistency by removing the information correlated with $\bfy$ from $\bfz$, so that any information pertaining to $\bfy$ that the decoder needs for reconstruction must all come from $\bfy$. While this model was previously used for image generation, we extend it into the sequential domain:
\eq{
\max_{\theta, \phi} \Bigg( \Expect_{q_{\phi}(\bfz|\tau)} \bigg[ \min_{\psi} \calL^{\text{aux}} \big( A_{\psi}(\bfz),\bfy \big) + \sum_{t=1}^T \log \pi_{\theta}(\bfa_t | \bfs_t, \bfz, \bfy) \bigg] - D_{KL} \big( q_{\theta}(\bfz|\tau) || p(\bfz) \big) \Bigg).
\label{eq:ctvae_info_obj_full}
}
Note that the encoder in \eqref{eq:ctvae_obj} and \eqref{eq:ctvae_info_obj_full} differ in that $q_{\phi}(\bfz|\tau)$ is no longer conditioned on the label $\bfy$.

\paragraph{3) CTVAE with mutual information maximization (CTVAE-mi).}
In addition to \eqref{eq:ctvae_obj}, we can also maximize the mutual information between labels and trajectories $I(\bfy;\tau)$. This quantity is hard to maximize directly, so instead we maximize the variational lower bound:
\eq{
I(\bfy;\tau) \geq \Expect_{\bfy \sim p(\bfy), \tau \sim \pi_{\theta}(\cdot | \bfz, \bfy)} \big[ \log r_{\psi}(\bfy|\tau) \big] + \calH(\bfy),
\label{eq:ctvae_mi_part}
}
where $r_{\psi}$ approximates the true posterior $p(\bfy|\tau)$. In our setting, the prior over labels is known, so $\calH(\bfy)$ is a constant. Thus, the learning objective is:
\eq{
\calL^{\text{ctvae-mi}}(\tau, \pi_{\theta}; q_{\phi}) = \calL^{\text{ctvae}}(\tau, \pi_{\theta}) + \Expect_{\bfy \sim p(\bfy), \tau \sim \pi_{\theta}(\cdot | \bfz, \bfy)} \big[ - \log r_{\psi}(\bfy|\tau) \big].
\label{eq:ctvae_mi_full}
}
Optimizing \eqref{eq:ctvae_mi_full} also requires collecting rollouts with the current policy, so similarly we also pretrain and fine-tune a dynamics model $M_{\varphi}$. This baseline can be interpreted as a supervised analogue of unsupervised models that maximize mutual information in \citep{li_17, hausman_17}.

\section{Stochastic Dynamics Function}
\label{app:stochastic_dynamics}

%
If the dynamics function $f$ of the environment is stochastic, we modify our approach in Algorithm \ref{alg:model-based-approach} by changing the form of our dynamics model. We can model the change in state as a Gaussian distribution and minimize the negative log-likelihood:
\eq{
\varphi_{\mu}^*, \varphi_{\sigma}^* = \argmin_{\varphi_{\mu}, \varphi_{\mu}} \Expect_{\tau \sim \calD} \sum_{t=1}^T - \log p(\Delta_t; \mu_t, \sigma_t),
\label{eq:dynamics_obj_stochastic}
}
where $\Delta_t = \bfs_{t+1} - \bfs_t$, $\mu_t = M_{\varphi_{\mu}}(\bfs_t, \bfa_t)$, $\sigma_t = M_{\varphi_{\sigma}}(\bfs_t, \bfa_t)$, and $M_{\varphi_{\mu}}$, $M_{\varphi_{\sigma}}$ are neural networks that can share weights. We can sample a change in state during rollouts using the reparametrization trick \citep{kingma2014auto}, which allows us to backpropagate through the dynamics model during training.

\section{Experiment Details}
\label{app:exp_details}

\begin{figure}[ht]
    \centering
    \begin{subfigure}[b]{0.19\columnwidth}
        \includegraphics[width=\columnwidth]{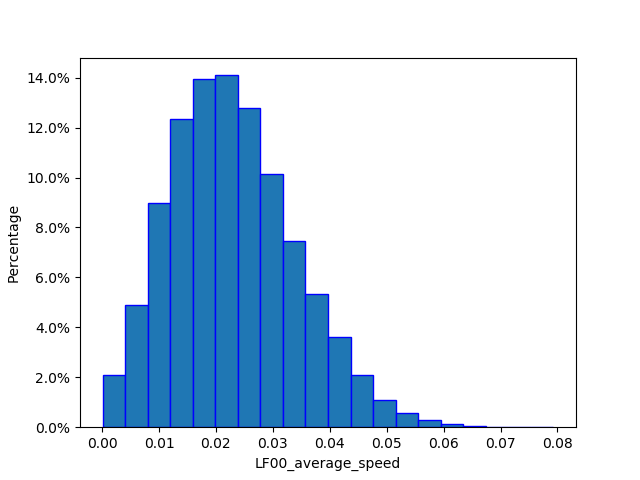}
        \caption{Speed}
    \end{subfigure}
    \begin{subfigure}[b]{0.19\columnwidth}
        \includegraphics[width=\columnwidth]{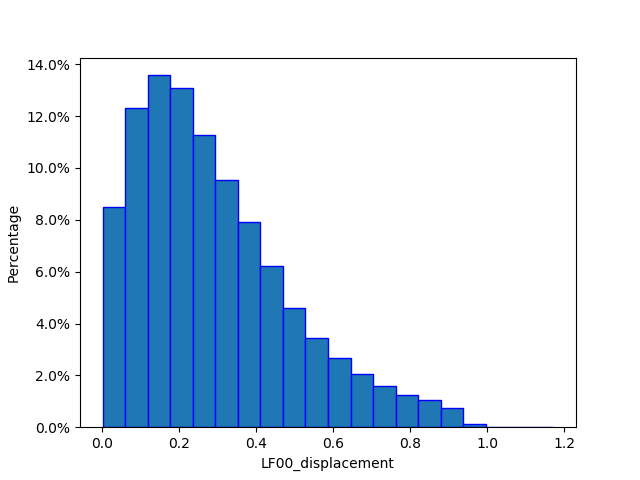}
        \caption{Displacement}
    \end{subfigure}
    \begin{subfigure}[b]{0.19\columnwidth}
        \includegraphics[width=\columnwidth]{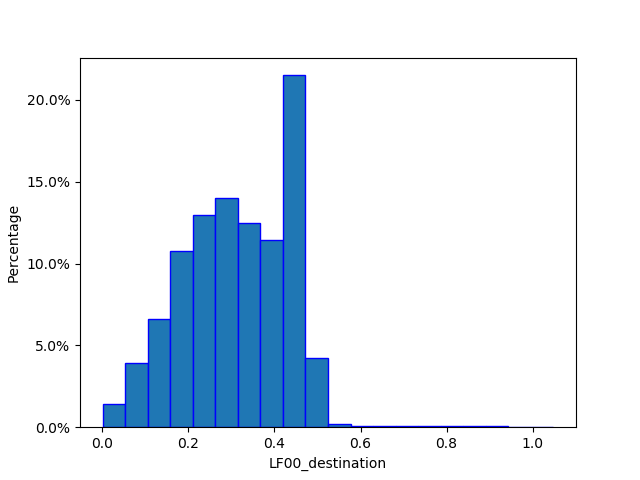}
        \caption{Destination}
    \end{subfigure}
    \begin{subfigure}[b]{0.19\columnwidth}
        \includegraphics[width=\columnwidth]{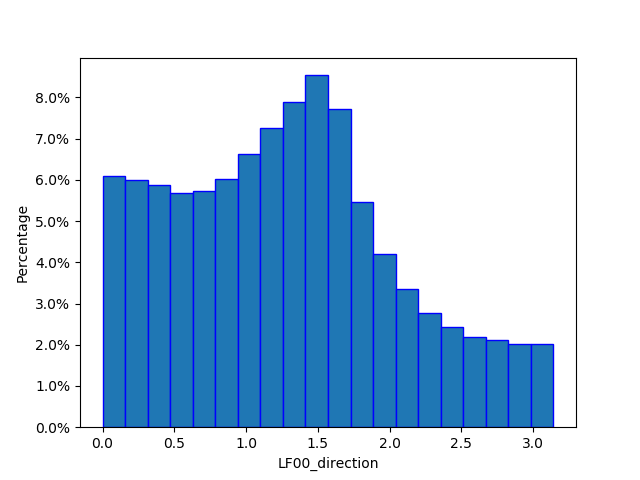}
        \caption{Direction}
    \end{subfigure}
    \begin{subfigure}[b]{0.19\columnwidth}
        \includegraphics[width=\columnwidth]{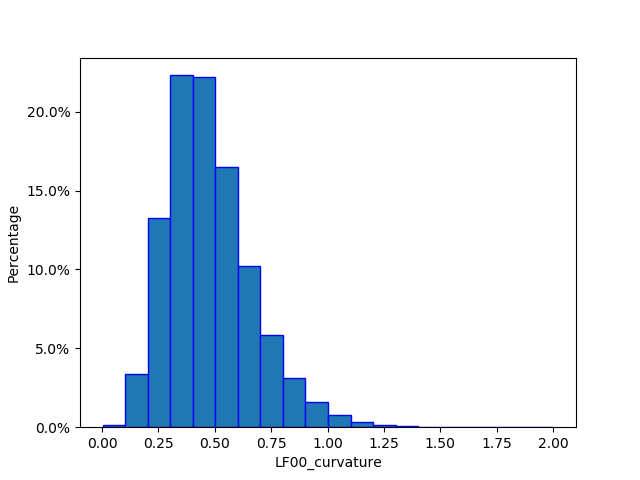}
        \caption{Curvature}
    \end{subfigure}
\caption{Histogram of basketball labeling functions applied on the training set (before applying thresholds). Basketball trajectories are collected from tracking real players in the NBA.}
\label{fig:bball_lf_hist}
\end{figure}

\begin{figure}[ht]
    \centering
    \begin{subfigure}[b]{0.24\columnwidth}
        \includegraphics[width=\columnwidth]{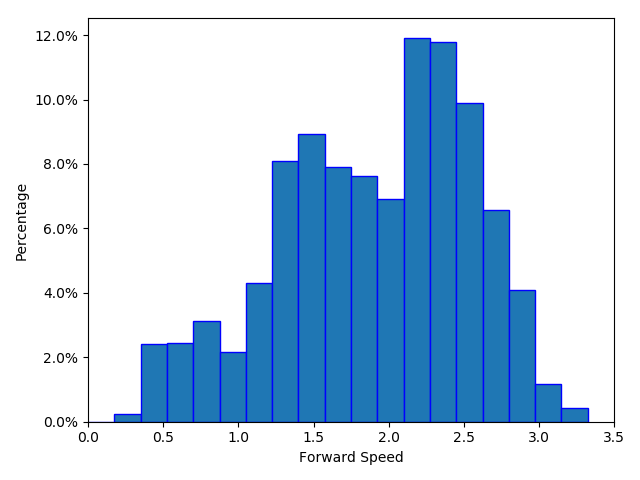}
        \caption{Speed}
    \end{subfigure}
    \begin{subfigure}[b]{0.24\columnwidth}
        \includegraphics[width=\columnwidth]{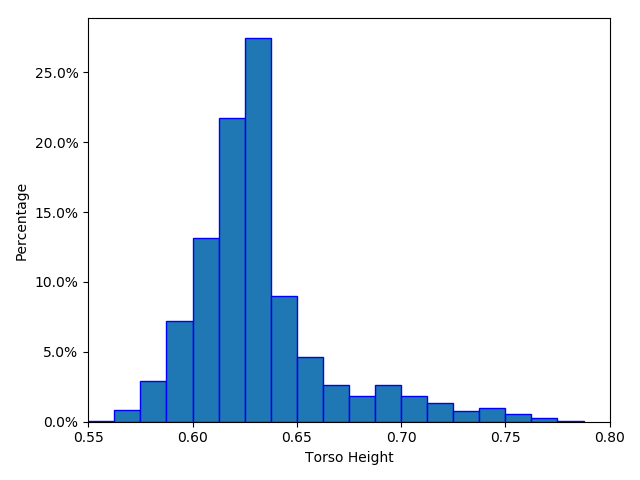}
        \caption{Torso Height}
    \end{subfigure}
    \begin{subfigure}[b]{0.24\columnwidth}
        \includegraphics[width=\columnwidth]{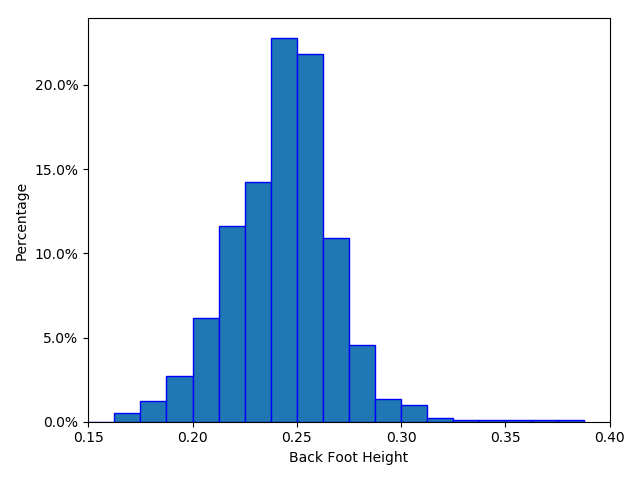}
        \caption{Back-Foot Height}
    \end{subfigure}
    \begin{subfigure}[b]{0.24\columnwidth}
        \includegraphics[width=\columnwidth]{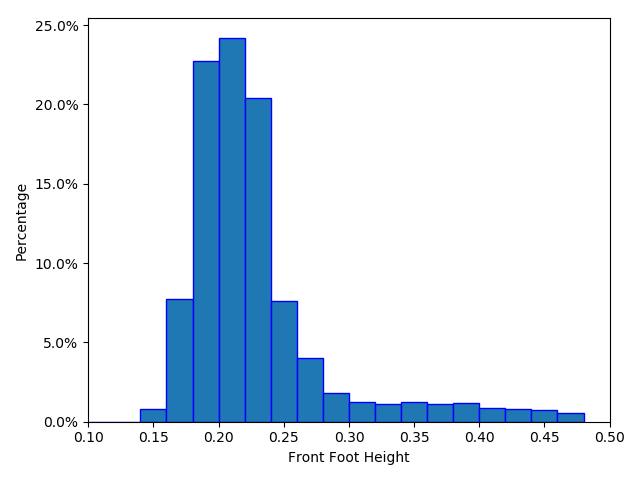}
        \caption{Front-Foot Height}
    \end{subfigure}
\caption{Histogram of Cheetah labeling functions applied on the training set (before applying thresholds). Note that \texttt{SPEED} is the most diverse behavior because we pre-trained the policies to achieve various speeds when collecting demonstrations, similar to \citep{wang_17}. For more diversity with respect to other behaviors, we can also incorporate a target behavior as part of the reward when pre-training Cheetah policies.}
\label{fig:cheetah_lf_hist}
\end{figure}

%
\paragraph{Dataset details.}
See Table \ref{tab:data_params}. Basketball trajectories are collected from tracking real players in the NBA. Figure \ref{fig:bball_lf_hist} shows the distribution of basketball labeling functions applied on the training set. For Cheetah, we train 125 policies using PPO \citep{ppo} to run forwards at speeds ranging from 0 to 4 (m/s). We collect 25 trajectories per policy by sampling actions from the policy. We use \citep{pytorchrl} to interface with \citep{dm_control}. Figure \ref{fig:cheetah_lf_hist} shows the distributions of Cheetah labeling functions applied on the training set.

\paragraph{Hyperparameters.} See Table \ref{tab:hyperparams} for training hyperparameters and Table \ref{tab:model_params} for model hyperparameters.

\begin{table}[ht!]
  \begin{center}
    \begin{tabular}{l|c|c|c|c|c|c}
      & $ |\calS|$ & $|\calA|$ & $T$ & $N_{\text{train}}$ & $N_{\text{test}}$ & frequency (Hz) \\
      \hline
      Basketball & 2  & 2 & 24  & 520,015 & 67,320 & 3  \\
      Cheetah    & 18 & 6 & 200 & 2,500   & 625    & 40 
    \end{tabular}
  \end{center}
  \caption{Dataset parameters for basketball and Cheetah environments.}
  \label{tab:data_params}
\end{table}

\begin{table}[ht!]
  \begin{center}
    \begin{tabular}{l|c|c|c|c|c|c|c|c}
       & batch size & \# batch $b$ & $n_{\text{dynamics}}$ & $n_{\text{label}}$ & $n_{\text{policy}}$ & $n_{\text{collect}}$ & $n_{\text{env}}$ & learning rate \\
      \hline
      Basketball & 128 & 4,063 & $10 \cdot b$ & $20 \cdot b$ & $30 \cdot b$ & 128 & 0 & $2\cdot10^{-4}$ \\
      Cheetah    & 16  & 157   & $50 \cdot b$ & $20 \cdot b$ & $60 \cdot b$ & 16  & 1  & $10^{-3}$
    \end{tabular}
  \end{center}
  \caption{Hyperparameters for Algorithm \ref{alg:model-based-approach}. $b$ is the number of batches to see all trajectories in the dataset once. We also use $L_2$ regularization of $10^{-5}$ for training the dynamics model $M_{\varphi}$.}
  \label{tab:hyperparams}
\end{table}

\begin{table}[ht!]
  \begin{center}
    \begin{tabular}{l|c|c|c|c|c|c}
      & $\bfz$-dim & $q_{\phi}$ GRU & $C_{\psi}^{\lambda}$ GRU & $\pi_{\theta}$ GRU & $\pi_{\theta}$ sizes & $M_{\varphi}$ sizes \\
      \hline
      Basketball & 4 & 128 & 128 & 128 & (128,128) & (128,128)  \\
      Cheetah    & 8 & 200 & 200 & -   & (200,200) & (500,500) 
    \end{tabular}
  \end{center}
  \caption{Model parameters for basketball and Cheetah environments.}
  \label{tab:model_params}
\end{table}

\section{Experiment Results}
\label{app:exp_results}


\begin{table}
    \centering
    \begin{subtable}{\linewidth}\centering
    {
        \begin{tabularx}{\textwidth}{ |l| *{15}{Y|} }
            \cline{1-16}
            \multicolumn{1}{|c|}{\tb{Model}} 
            & \multicolumn{3}{c|}{\tb{Speed}}
            & \multicolumn{3}{c|}{\tb{Displacement}}
            & \multicolumn{3}{c|}{\tb{Destination}}
            & \multicolumn{3}{c|}{\tb{Direction}}
            & \multicolumn{3}{c|}{\tb{Curvature}} \\
            \hline
            CTVAE      & 82 & 83 & 85 & 71 & 72 & 74 & 81 & 82 & 82 & 76 & 77 & 80 & 60 & 61 & 62 \\
            CTVAE-info & \tb{84} & 84 & 87 & 69 & 71 & 74 & 78 & 79 & 83 & 71 & 72 & 74 & 60 & 60 & 62 \\
            CTVAE-mi   & 84 & 86 & 87 & 71 & 74 & 74 & 80 & 82 & 84 & 75 & 77 & 78 & 58 & 72 & 74 \\
            \hline
            CTVAE-style & 34 & \tb{95} & \tb{97} & \tb{89} & \tb{96} & \tb{97} & \tb{91} & \tb{97} & \tb{98} & \tb{96} & \tb{97} & \tb{98} & \tb{77} & \tb{81} & \tb{83} \\
            \hline
        \end{tabularx}
    }
    \caption{\Metric{} wrt. single styles of 3 classes (roughly uniform distributions).}
    \vspace{4pt}
    \label{tab:bball_exp1}
    \end{subtable}
    \begin{subtable}{\linewidth}\centering
    {
        \begin{tabularx}{\textwidth}{ |l| *{15}{Y|} }
            \cline{1-16}
            \multicolumn{1}{|c|}{\tb{Model}} 
            & \multicolumn{3}{c|}{\tb{2 classes}}
            & \multicolumn{3}{c|}{\tb{3 classes}}
            & \multicolumn{3}{c|}{\tb{4 classes}}
            & \multicolumn{3}{c|}{\tb{6 classes}}
            & \multicolumn{3}{c|}{\tb{8 classes}} \\
            \hline
            CTVAE       & 91 & 92 & 93 & 79 & 83 & 84 & \tb{76} & 79 & 79 & \tb{68} & 70 & 72 & 64 & 66 & 69 \\
            CTVAE-info  & 90 & 90 & 92 & \tb{83} & 83 & 85 & 75 & 76 & 77 & 68 & 70 & 72 & 60 & 63 & 67 \\
            CTVAE-mi    & 90 & 92 & 93 & 81 & 84 & 86 & 75 & 77 & 80 & 66 & 70 & 72 & 62 & 62 & 67 \\
            \hline
            CTVAE-style & \tb{98} & \tb{99} & \tb{99} & 15 & \tb{98} & \tb{99} & 15 & \tb{96} & \tb{96} & 02 & \tb{92} & \tb{94} & \tb{80} & \tb{90} & \tb{93} \\
            \hline
        \end{tabularx}
    }
    \caption{\Metric{} wrt. \texttt{DISPLACEMENT} of up to 8 classes (roughly uniform distributions).}
    \vspace{4pt}
    \label{tab:bball_exp2b}
    \end{subtable}
    \begin{subtable}{0.85\linewidth}\centering
    {
        \begin{tabularx}{\textwidth}{ |l| *{12}{Y|} }
            \cline{1-13}
            \multicolumn{1}{|c|}{\tb{Model}} 
            & \multicolumn{3}{c|}{\tb{2 classes}}
            & \multicolumn{3}{c|}{\tb{3 classes}}
            & \multicolumn{3}{c|}{\tb{4 classes}}
            & \multicolumn{3}{c|}{\tb{6 classes}} \\
            \hline
            CTVAE      & 86 & 87 & 87 & 80 & 82 & 83 & \tb{76} & 78 & 79 & 70 & 74 & 77 \\
            CTVAE-info & 83 & 87 & 88 & 79 & 81 & 83 & 73 & 75 & 78 & 71 & 77 & 78 \\
            CTVAE-mi   & 86 & 88 & 88 & \tb{80} & 81 & 84 & 71 & 74 & 79 & \tb{73} & 76 & 78 \\
            \hline
            CTVAE-style   & \tb{97} & \tb{98} & \tb{99} & 68 & \tb{97} & \tb{98} & 35 & \tb{89} & \tb{95} & 67 & \tb{84} & \tb{93} \\
            \hline
        \end{tabularx}
    }
    \caption{\Metric{} wrt. \texttt{DESTINATION(net)} with up to 6 classes (non-uniform distributions).}
    \vspace{4pt}
    \label{tab:bball_exp2}
    \end{subtable}
    \begin{subtable}{\linewidth}\centering
    {
        \begin{tabularx}{\textwidth}{ |l| *{15}{Y|} }
        \cline{1-16}
        \multicolumn{1}{|c|}{\tb{Model}}
        & \multicolumn{3}{c|}{\tb{2 styles 3 classes}}
        & \multicolumn{3}{c|}{\tb{3 styles 3 classes}}
        & \multicolumn{3}{c|}{\tb{4 styles 3 classes}}
        & \multicolumn{3}{c|}{\tb{5 styles 3 classes}}
        & \multicolumn{3}{c|}{\tb{5 styles 4 classes}} \\
        \hline
        CTVAE      & 67 & 71 & 73 & 58 & 58 & 62 & 49 & 50 & 52 & 27 & 37 & 35 & \tb{20} & 21 & 22 \\
        CTVAE-info & 68 & 69 & 70 & 54 & 58 & 59 & 48 & 51 & 54 & 28 & 32 & 35 & 18 & 21 & 23 \\
        CTVAE-mi   & 71 & 72 & 73 & 48 & 56 & 61 & 45 & 51 & 52 & 16 & 30 & 31 & 18 & 21 & 23 \\
        \hline
        CTVAE-style   & \tb{92} & \tb{93} & \tb{94} & \tb{86} & \tb{88} & \tb{90} & \tb{62} & \tb{88} & \tb{88} & \tb{66} & \tb{75} & \tb{80} & 11 & \tb{55} & \tb{77} \\
        \hline
    \end{tabularx}
    }
    \caption{\Metric{} wrt. multiple styles simultaneously.}
    \vspace{-4pt}
    \label{tab:bball_exp3}
    \end{subtable}
\caption{[min, median, max] \metric{} ($\times 10^{-2}$, 5 seeds) of policies evaluated with 4,000 basketball rollouts each. CTVAE-style policies significantly outperform baselines in all experiments and are calibrated at almost maximal \metric{} for 4/5 labeling functions. We note some rare failure cases with our approach, which we leave as a direction for improvement for future work.}
\label{tab:bball_exp_all}
\vspace{-10pt}
\end{table}

\begin{table}
    \centering
    \begin{subtable}{0.85\linewidth}\centering
    {
        \begin{tabularx}{\textwidth}{ |l| *{12}{Y|} }
            \cline{1-13}
            \multicolumn{1}{|c|}{\tb{Model}} 
            & \multicolumn{3}{c|}{\tb{Speed}}
            & \multicolumn{3}{c|}{\tb{Torso Height}}
            & \multicolumn{3}{c|}{\tb{B-Foot Height}}
            & \multicolumn{3}{c|}{\tb{F-Foot Height}} \\
            \hline
            CTVAE       & 53 & 59 & 62 & 62 & 63 & 70 & 61 & 68 & 73 & 63 & 68 & 72 \\
            CTVAE-info  & 56 & 57 & 61 & 62 & 63 & 72 & 58 & 65 & 72 & 63 & 66 & 69 \\
            CTVAE-mi    & 53 & 60 & 62 & 62 & 65 & 70 & 60 & 65 & 70 & 66 & 70 & 73 \\
            \hline
            CTVAE-style & \tb{68} & \tb{79} & \tb{81} & \tb{79} & \tb{80} & \tb{84} & \tb{77} & \tb{80} & \tb{88} & \tb{74} & \tb{77} & \tb{80} \\
            \hline
        \end{tabularx}
    }
    \caption{\Metric{} wrt. single styles of 2 classes (roughly uniform distributions).}
    \label{tab:cheetah_exp1}
    \vspace{4pt}
    \end{subtable}
    \begin{subtable}[t]{0.49\linewidth}
    {
        \begin{tabularx}{\textwidth}{ |l| *{6}{Y|} }
            \cline{1-7}
            \multicolumn{1}{|c|}{\tb{Model}}
            & \multicolumn{3}{c|}{\tb{3 classes}}
            & \multicolumn{3}{c|}{\tb{4 classes}} \\
            \hline
            CTVAE       & 41 & 45 & 49 & 35 & 37 & 41 \\
            CTVAE-info  & 47 & 49 & 52 & 36 & 39 & 42 \\
            CTVAE-mi    & 47 & 48 & 53 & 36 & 37 & 38 \\
            \hline
            CTVAE-style & \tb{59} & \tb{59} & \tb{65} & \tb{42} & \tb{51} & \tb{60} \\
            \hline
        \end{tabularx}
    }
    \caption{\Metric{} wrt. \texttt{SPEED} with varying \# of classes (non-uniform distributions).}
    \label{tab:cheetah_exp2}
    \end{subtable}
    \hfill
    \begin{subtable}[t]{0.49\linewidth}
    {
        \begin{tabularx}{\textwidth}{ |l| *{6}{Y|} }
            \cline{1-7}
            \multicolumn{1}{|c|}{\tb{Model}}
            & \multicolumn{3}{c|}{\tb{2 styles 2 classe}}
            & \multicolumn{3}{c|}{\tb{3 styles 2 classes}} \\
            \hline
            CTVAE       & 39 & 41 & 43 & 25 & 28 & 29 \\
            CTVAE-info  & 39 & 41 & 46 & 25 & 27 & 30 \\
            CTVAE-mi    & 34 & 40 & 48 & 27 & 28 & 31 \\
            \hline
            CTVAE-style & \tb{43} & \tb{54} & \tb{60} & \tb{38} & \tb{40} & \tb{52} \\
            \hline
        \end{tabularx}
    }
    \caption{\Metric{} wrt. multiple styles simultaneously.}
    \label{tab:cheetah_exp3}
    \end{subtable}
\caption{[min, median, max] \metric{} ($\times 10^{-2}$, 5 seeds) of policies evaluated with 500 Cheetah rollouts each. CTVAE-style policies consistently outperform all baselines, but we note that there is still room for improvement (to reach 100\% \metric{}).}
\label{tab:cheetah_exp_all}
\vspace{-10pt}
\end{table}

\begin{table}
    \centering
    \begin{subtable}{\linewidth}\centering
    {
        \begin{tabularx}{0.85\textwidth}{ |l| *{5}{Y|} }
            \cline{1-6}
            \multicolumn{1}{|c|}{\tb{Model}} 
            & \multicolumn{1}{c|}{\tb{Speed}}
            & \multicolumn{1}{c|}{\tb{Displacement}}
            & \multicolumn{1}{c|}{\tb{Destination}}
            & \multicolumn{1}{c|}{\tb{Direction}}
            & \multicolumn{1}{c|}{\tb{Curvature}} \\
            \hline
            CTVAE       & 83.4 $\pm$ 1.2 & 72.4 $\pm$ \tb{1.4} & 81.9 $\pm$ \tb{0.6} & 77.7 $\pm$ 1.3 & 61.0 $\pm$ 1.0  \\
            CTVAE-info  & 85.0 $\pm$ \tb{1.2} & 71.2 $\pm$ 1.9 & 80.1 $\pm$ 1.8 & 72.3 $\pm$ 1.1 & 60.2 $\pm$ \tb{0.8}  \\
            CTVAE-mi    & \tb{85.8} $\pm$ 1.3 & 72.8 $\pm$ 1.5 & 82.2 $\pm$ 1.4 & 76.9 $\pm$ 1.1 & 68.6 $\pm$ 6.4  \\
            \hline
            CTVAE-style & 72.1 $\pm$ 33.3 & \tb{94.6} $\pm$ 3.1 & \tb{95.0} $\pm$ 3.7 & \tb{96.8} $\pm$ \tb{0.7} & \tb{79.6} $\pm$ 2.7  \\
            \hline
        \end{tabularx}
    }
    \caption{\Metric{} wrt. single styles of 3 classes (roughly uniform distributions).}
    \vspace{4pt}
    \label{tab:bball_exp1_meanstd}
    \end{subtable}
    \begin{subtable}{0.85\linewidth}\centering
    {
        \begin{tabularx}{\textwidth}{ |l| *{5}{Y|} }
            \cline{1-6}
            \multicolumn{1}{|c|}{\tb{Model}} 
            & \multicolumn{1}{c|}{\tb{2 classes}}
            & \multicolumn{1}{c|}{\tb{3 classes}}
            & \multicolumn{1}{c|}{\tb{4 classes}}
            & \multicolumn{1}{c|}{\tb{6 classes}}
            & \multicolumn{1}{c|}{\tb{8 classes}} \\
            \hline
            CTVAE       & 92.1 $\pm$ 0.9 & 82.4 $\pm$ 2.4 & 78.0 $\pm$ 1.4 & 69.9 $\pm$ \tb{1.4} & 66.0 $\pm$ \tb{2.0} \\
            CTVAE-info  & 90.5 $\pm$ 0.9 & \tb{83.6} $\pm$ \tb{1.0} & 75.9 $\pm$ \tb{0.9} & \tb{70.2} $\pm$ 1.6 & 63.4 $\pm$ 2.9 \\
            CTVAE-mi    & 91.6 $\pm$ 1.2 & 83.5 $\pm$ 2.1 & 77.6 $\pm$ 2.5 & 68.8 $\pm$ 2.5 & 63.7 $\pm$ 2.3 \\
            \hline
            CTVAE-style & \tb{98.7} $\pm$ \tb{0.4} & 81.4 $\pm$ 36.9 & \tb{79.3} $\pm$ 35.9 & 68.1 $\pm$ 40.0 & \tb{88.2} $\pm$ 5.1 \\
            \hline
        \end{tabularx}
    }
    \caption{\Metric{} wrt. \texttt{DISPLACEMENT} of up to 8 classes (non-uniform distributions).}
    \vspace{4pt}
    \label{tab:bball_exp2b_meanstd}
    \end{subtable}
    \begin{subtable}{0.7\linewidth}\centering
    {
        \begin{tabularx}{\textwidth}{ |l| *{4}{Y|} }
            \cline{1-5}
            \multicolumn{1}{|c|}{\tb{Model}} 
            & \multicolumn{1}{c|}{\tb{2 classes}}
            & \multicolumn{1}{c|}{\tb{3 classes}}
            & \multicolumn{1}{c|}{\tb{4 classes}}
            & \multicolumn{1}{c|}{\tb{6 classes}} \\
            \hline
            CTVAE      & 86.6 $\pm$ \tb{0.6} & 81.6 $\pm$ \tb{1.3} & \tb{77.4} $\pm$ \tb{1.5} & 74.0 $\pm$ 2.6 \\
            CTVAE-info & 86.2 $\pm$ 1.7 & 81.1 $\pm$ 1.4 & 75.3 $\pm$ 2.5 & 75.3 $\pm$ 3.3 \\
            CTVAE-mi   & 87.3 $\pm$ 0.9 & 81.6 $\pm$ 1.6 & 74.3 $\pm$ 3.1 & 75.8 $\pm$ \tb{2.1} \\
            \hline
            CTVAE-style & \tb{98.1} $\pm$ 0.8 & \tb{88.2} $\pm$ 13.6 & 77.0 $\pm$ 24.1 & \tb{82.6} $\pm$ 11.3 \\
            \hline
        \end{tabularx}
    }
    \caption{\Metric{} wrt. \texttt{DESTINATION(net)} of up to 6 classes (non-uniform distributions).}
    \vspace{4pt}
    \label{tab:bball_exp2_meanstd}
    \end{subtable}
    \begin{subtable}{\linewidth}\centering
    {
        \begin{tabularx}{\textwidth}{ |l| *{5}{Y|} }
        \cline{1-6}
        \multicolumn{1}{|c|}{\tb{Model}}
        & \multicolumn{1}{c|}{\tb{2 styles 3 classes}}
        & \multicolumn{1}{c|}{\tb{3 styles 3 classes}}
        & \multicolumn{1}{c|}{\tb{4 styles 3 classes}}
        & \multicolumn{1}{c|}{\tb{5 styles 3 classes}}
        & \multicolumn{1}{c|}{\tb{5 styles 4 classes}} \\
        \hline
        CTVAE      & 70.5 $\pm$ 2.1 & 58.9 $\pm$ \tb{1.5} & 50.4 $\pm$ \tb{1.4} & 31.6 $\pm$ 2.8 & \tb{20.8} $\pm$ 1.0 \\
        CTVAE-info & 69.0 $\pm$ 0.9 & 57.5 $\pm$ 2.0 & 50.5 $\pm$ 2.3 & 31.4 $\pm$ \tb{2.5} & 20.6 $\pm$ 2.0 \\
        CTVAE-mi   & 71.8 $\pm$ \tb{0.7} & 53.8 $\pm$ 5.9 & 50.2 $\pm$ 2.7 & 26.9 $\pm$ 6.3 & 20.7 $\pm$ 1.9 \\
        \hline
        CTVAE-style & \tb{92.8} $\pm$ 1.0 & \tb{88.3} $\pm$ 1.7 & \tb{81.7} $\pm$ 11.0 & \tb{73.9} $\pm$ 5.4 & \tb{50.3} $\pm$ 24.7 \\
        \hline
    \end{tabularx}
    }
    \caption{\Metric{} wrt. multiple styles simultaneously.}
    \vspace{-4pt}
    \label{tab:bball_exp3_meanstd}
    \end{subtable}
\caption{Mean and standard deviation \metric{} ($\times 10^{-2}$, 5 seeds) of policies evaluated with 4,000 basketball rollouts each. CTVAE-style policies generally outperform baselines. Lower mean \metric{} (and large standard deviation) for CTVAE-style is often due to failure cases, as can be seen from the minimum \metric{} values we report in Table \ref{tab:bball_exp_all}. Understanding the causes of these failure cases and improving the algorithm's stability are possible directions for future work.}
\label{tab:bball_exp_all_meanstd}
\vspace{-10pt}
\end{table}

\begin{table}
    \centering
    \begin{subtable}{\linewidth}\centering
    {
        \begin{tabularx}{0.75\textwidth}{ |l| *{4}{Y|} }
            \cline{1-5}
            \multicolumn{1}{|c|}{\tb{Model}} 
            & \multicolumn{1}{c|}{\tb{Speed}}
            & \multicolumn{1}{c|}{\tb{Torso Height}}
            & \multicolumn{1}{c|}{\tb{B-Foot Height}}
            & \multicolumn{1}{c|}{\tb{F-Foot Height}} \\
            \hline
            CTVAE       & 57.4 $\pm$ 3.9 & 64.4 $\pm$ 3.1 & 67.4 $\pm$ 4.2 & 68.5 $\pm$ 3.7 \\
            CTVAE-info  & 58.3 $\pm$ \tb{2.1} & 65.0 $\pm$ 4.2 & 64.1 $\pm$ 5.4 & 66.1 $\pm$ 2.7 \\
            CTVAE-mi    & 58.4 $\pm$ 3.9 & 65.7 $\pm$ 3.2 & 65.0 $\pm$ \tb{3.6} & 69.9 $\pm$ 2.6 \\
            \hline
            CTVAE-style & \tb{77.0} $\pm$ 5.3 & \tb{81.0} $\pm$ \tb{2.2} & \tb{81.9} $\pm$ 5.4 & \tb{77.2} $\pm$ \tb{2.4} \\
            \hline
        \end{tabularx}
    }
    \caption{\Metric{} wrt. single styles of 2 classes (roughly uniform distributions).}
    \label{tab:cheetah_exp1_meanstd}
    \vspace{4pt}
    \end{subtable}
    \begin{subtable}[t]{0.49\linewidth}
    {
        \begin{tabularx}{\textwidth}{ |l| *{2}{Y|} }
            \cline{1-3}
            \multicolumn{1}{|c|}{\tb{Model}}
            & \multicolumn{1}{c|}{\tb{3 classes}}
            & \multicolumn{1}{c|}{\tb{4 classes}} \\
            \hline
            CTVAE       & 45.2 $\pm$ 3.2 & 37.8 $\pm$ 2.9 \\
            CTVAE-info  & 49.2 $\pm$ \tb{1.8} & 39.3 $\pm$ 2.8 \\
            CTVAE-mi    & 49.1 $\pm$ 2.2 & 36.8 $\pm$ \tb{1.0} \\
            \hline
            CTVAE-style & \tb{60.8} $\pm$ 2.9 & \tb{51.3} $\pm$ 7.8 \\
            \hline
        \end{tabularx}
    }
    \caption{\Metric{} wrt. \texttt{SPEED} with varying \# of classes (non-uniform distributions).}
    \label{tab:cheetah_exp2_meanstd}
    \end{subtable}
    \hfill
    \begin{subtable}[t]{0.49\linewidth}
    {
        \begin{tabularx}{\textwidth}{ |l| *{2}{Y|} }
        \cline{1-3}
        \multicolumn{1}{|c|}{\tb{Model}}
        & \multicolumn{1}{c|}{\tb{2 styles 2 classes}}
        & \multicolumn{1}{c|}{\tb{3 styles 2 classes}} \\
        \hline
        CTVAE       & 40.9 $\pm$ \tb{1.6} & 27.2 $\pm$ 1.9 \\
        CTVAE-info  & 41.8 $\pm$ 2.3 & 27.8 $\pm$ 2.2 \\
        CTVAE-mi    & 40.7 $\pm$ 4.9 & 28.5 $\pm$ \tb{1.6} \\
        \hline
        CTVAE-style & \tb{52.6} $\pm$ 6.1 & \tb{42.8} $\pm$ 5.8 \\
        \hline
    \end{tabularx}
    }
    \caption{\Metric{} wrt. multiple styles simultaneously.}
    \label{tab:cheetah_exp3_meanstd}
    \end{subtable}
\caption{Mean and standard deviation \metric{} ($\times 10^{-2}$, 5 seeds) of policies evaluated with 500 Cheetah rollouts each. CTVAE-style policies consistently outperform all baselines, but we note that there is still room for improvement (to reach 100\% \metric{}).}
\label{tab:cheetah_exp_all_meanstd}
\vspace{-10pt}
\end{table}

\begin{table}[ht!]
\centering
    \begin{tabularx}{0.8\textwidth}{ |l| *{8}{Y|} }
        \cline{2-9}
        \multicolumn{1}{c|}{} 
        & \multicolumn{2}{c|}{\tb{Speed}}
        & \multicolumn{2}{c|}{\tb{Torso Height}}
        & \multicolumn{2}{c|}{\tb{B-Foot Height}}
        & \multicolumn{2}{c|}{\tb{F-Foot Height}} \\
        \hline
        \tb{Model}   & NLD & SC & NLD & SC & NLD & SC & NLD & SC \\
        \hline
        CTVAE-style  & -0.28 & \tb{79} & -0.24 & 80 & -0.16 & \tb{80} & -0.22 & \tb{77} \\
        CTVAE-style+ & -0.49 & 70 & -0.42 & \tb{83} & -0.36 & \tb{80} & -0.42 & 74 \\
        \hline
    \end{tabularx}
\caption{We report the median negative log-density per timestep (lower is better) and \metric{} (higher is better) of CTVAE-style policies for Cheetah (5 seeds). The first row corresponds to experiments in Tables \ref{tab:exp_1} and \ref{tab:cheetah_exp1}, and the second row corresponds to the same experiments with 50\% more training iterations. The KL-divergence in the two sets of experiments are roughly the same. Although imitation quality improves, \metric{} can sometimes degrade (e.g. \texttt{SPEED}, \texttt{FRONT-FOOT HEIGHT}), indicating a possible trade-off between imitation quality and \metric{}.}
\label{tab:cheetah_tradeoff}
\end{table}

\begin{table}[h]
\centering
    \begin{tabularx}{0.65\columnwidth}{ |l| *{6}{Y|} }
        \cline{2-6}
        \multicolumn{1}{c|}{} 
        & \multicolumn{5}{c|}{\tb{\Metric{} $\uparrow$}}
        & \multicolumn{1}{c}{} \\
        \hline
        \tb{Model} & \tb{Min} & - & \tb{Median} & - & \tb{Max} & NLD $\downarrow$ \\
        \hline
        RNN        & 79 & 79 & 80 & 81 & 81 & -7.7 \\
        RNN-style  & 81 & 86 & 91 & 95 & 98 & -7.6 \\
        \hline
        CTVAE       & 81 & 82 & 82 & 82 & 82 & \tb{-8.0} \\
        CTVAE-style & \tb{91} & \tb{92} & \tb{97} & \tb{98} & \tb{98} & -7.8 \\
        \hline
    \end{tabularx}
    \vspace{-0.1in}
\caption{Comparing \metric{} ($\times 10^{-2}$) between RNN and CTVAE policy models for \texttt{DESTINATION} in basketball. The \metric{} for 5 seeds are listed in increasing order. Our algorithm improves \metric{} for both policy models at the cost of a slight degradation in imitation quality. In general, CTVAE performs better than RNN in both \metric{} and imitation quality.}
\label{tab:rnn_exp}
\end{table}

\begin{table}
    \begin{subtable}{\linewidth}\centering
    {
        \begin{tabularx}{0.8\textwidth}{ |l| *{5}{Y|} }
            \cline{2-6}
            \multicolumn{1}{c|}{} 
            & \multicolumn{1}{c|}{\tb{Speed}}
            & \multicolumn{1}{c|}{\tb{Displacement}}
            & \multicolumn{1}{c|}{\tb{Destination}}
            & \multicolumn{1}{c|}{\tb{Direction}}
            & \multicolumn{1}{c|}{\tb{Curvature}} \\
            \hline
            $\calL^{\text{label}}$ & 3.96 $\pm$ 0.33 & 4.58 $\pm$ 0.20 & 1.61 $\pm$ 0.18 & 3.19 $\pm$ 0.25 & 28.31 $\pm$ 0.95  \\
            \hline
        \end{tabularx}
    }
    \caption{Basketball labeling functions for experiments in section \ref{sec:exp1}.}
    \vspace{4pt}
    \label{tab:bball_C_approx}
    \end{subtable}
    \begin{subtable}{\linewidth}\centering
    {
        \begin{tabularx}{0.7\textwidth}{ |l| *{4}{Y|} }
            \cline{2-5}
            \multicolumn{1}{c|}{} 
            & \multicolumn{1}{c|}{\tb{Speed}}
            & \multicolumn{1}{c|}{\tb{Torso Height}}
            & \multicolumn{1}{c|}{\tb{B-Foot Height}}
            & \multicolumn{1}{c|}{\tb{F-Foot Height}} \\
            \hline
            $\calL^{\text{label}}$ & 3.24 $\pm$ 0.83 & 15.87 $\pm$ 1.78 & 17.25 $\pm$ 0.73 & 14.75 $\pm$ 0.74 \\
            \hline
        \end{tabularx}
    }
    \caption{Cheetah labeling functions for experiments in section \ref{sec:exp1}.}
    \vspace{4pt}
    \label{tab:cheetah_C_approx}
    \end{subtable}
\caption{Mean and standard deviation cross-entropy loss ($\calL^{\text{label}}, \times 10^{-2}$) over 5 seeds of learned label approximators $C_{\psi^*}^{\lambda}$ on test trajectories after $n^{\text{label}}$ training iterations for experiments in section \ref{sec:exp1}. $C_{\psi^*}^{\lambda}$ is only used during training; when computing \metric{} for our quantitative results, we use original labeling functions $\lambda$.}
\label{tab:C_approx}
\vspace{-10pt}
\end{table}

\begin{table}[ht!]
  \begin{center}
    \begin{tabular}{l|c}
      & $M_{\varphi}$ test error \\
      \hline
      Basketball & $1.47 \pm 0.59 (\times 10^{-7})$ \\
      Cheetah    & $1.93 \pm 0.08 (\times 10^{-2})$
    \end{tabular}
  \end{center}
  \caption{Average mean-squared error of dynamics model $M_{\varphi}$ per timestep per dimension on test data after training for $n^{\text{dynamics}}$ iterations}
  \label{tab:dynamics_performance}
\end{table}

\begin{table}
\centering
    \begin{tabularx}{0.6\linewidth}{ |c| *{5}{Y|} }
        \cline{2-6}
        \multicolumn{1}{c|}{} 
        & \multicolumn{5}{c|}{\tb{Basketball}} \\
        \hline
        \tb{noise} & \tb{Speed} & \tb{Disp.} & \tb{Dest.} & \tb{Dir.} & \tb{Curve} \\
        \hline
        $\sigma$  & 5.20  & 6.18  & 7.46  & 5.36  & 5.88  \\
        2$\sigma$ & 10.33 & 12.24 & 15.54 & 10.93 & 11.66 \\
        3$\sigma$ & 15.36 & 18.08 & 23.46 & 16.78 & 17.24 \\
        4$\sigma$ & 20.10 & 23.47 & 30.10 & 22.56 & 22.52 \\
        \hline
        $\sigma$ \text{value} & 0.001 & 0.02 & 0.02 & 0.1 & 0.02 \\
        \hline
    \end{tabularx}
\caption{\textbf{Label disagreement ($\%$) of noisy labeling functions:} For each of the Basketball labeling functions with 3 classes in Table \ref{tab:exp_1}, we consider noisy versions where we inject Gaussian noise with mean 0 and standard deviation $c \cdot \sigma$ for $c \in \{ 1, 2, 3, 4 \}$ before applying thresholds to obtain label classes. This table shows the label disagreement between noisy and true labeling functions over trajectories in the training set. The last row shows the $\sigma$ value used for each labeling function.}
\label{tab:exp_noise_mismatch}
\end{table}

\begin{table}
\centering
    \begin{tabularx}{0.6\linewidth}{ |c| *{5}{Y|} }
        \cline{2-6}
        \multicolumn{1}{c|}{} 
        & \multicolumn{5}{c|}{\tb{Basketball}} \\
        \hline
        \tb{noise} & \tb{Speed} & \tb{Disp.} & \tb{Dest.} & \tb{Dir.} & \tb{Curve} \\
        \hline
        $\sigma$  & 2.78  & 3.21  & 3.70  & 3.71  & 3.16  \\
        2$\sigma$ & 5.59  & 7.88  & 9.75  & 8.63  & 4.46  \\
        3$\sigma$ & 9.71  & 15.37 & 16.38 & 12.39 & 6.34  \\
        4$\sigma$ & 11.63 & 20.54 & 21.11 & 19.98 & 12.41 \\
        \hline
    \end{tabularx}
\caption{\textbf{Relative \textit{decrease} in \metric{}} when training with noisy labeling functions: ($\%$, median over 5 seeds) Using the noisy labeling functions in Table \ref{tab:exp_noise_mismatch}, we train CTVAE-style models and evaluate \metric{} using the true labeling functions without noise. This table shows the percentage decrease in \metric{} relative to when there is no noise in Table \ref{tab:exp_1}. Comparing with the label disagreement in Table \ref{tab:exp_noise_mismatch}, we see that the relative decrease in \metric{} generally scales linearly with the label disagreement between noisy and true labeling functions.}
\label{tab:exp_noise}
\end{table}


\begin{figure*}
  \begin{minipage}[c]{0.7\textwidth}
    \begin{center}
        \begin{figure}[H]
        \begin{center}
        \begin{subfigure}[t]{0.325\columnwidth}
        \centering
        \includegraphics[width=.8\columnwidth]{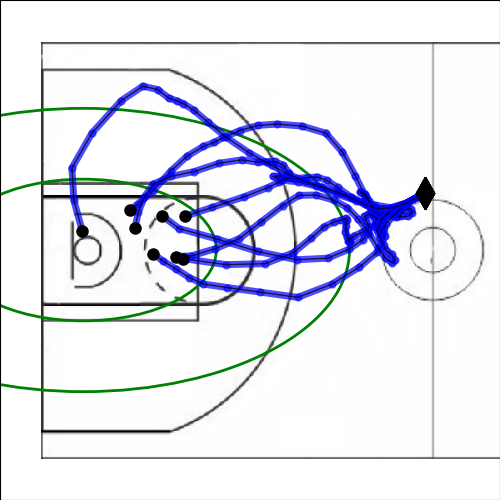}
        \caption{Label class 0 (close)}
        \label{fig:bball_lf_dest_0}
        \end{subfigure}
        \begin{subfigure}[t]{0.325\columnwidth}
        \centering
        \includegraphics[width=.8\columnwidth]{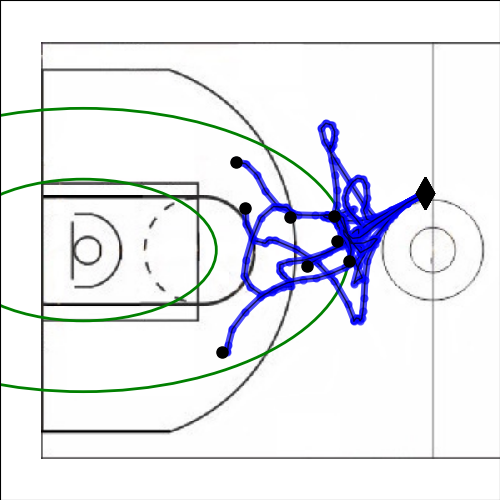}
        \caption{Label class 1 (mid)}
        \label{fig:bball_lf_dest_1}
        \end{subfigure}
        \begin{subfigure}[t]{0.325\columnwidth}
        \centering
        \includegraphics[width=.8\columnwidth]{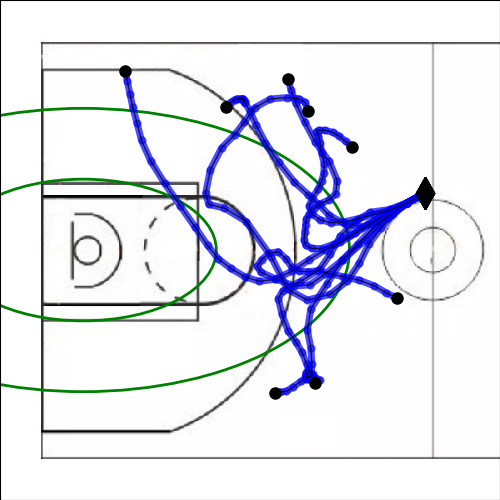}
        \caption{Label class 2 (far)}
        \label{fig:bball_lf_dest_2}
        \end{subfigure}
        \end{center}
        \end{figure}
    \end{center}
  \end{minipage}\hfill
  \begin{minipage}[c]{0.29\textwidth}
    \caption{CTVAE-style rollouts calibrated for \texttt{DESTINATION(net)}, 0.97 \metric{}. Diamonds ($\blacklozenge$) and dots ($\bullet$) indicate initial and final positions. Regions divided by green lines represent label classes.}
    \label{fig:bball_lf_dest}
  \end{minipage}
\end{figure*}

\begin{figure}[ht]
    \centering
    \begin{subfigure}[b]{0.32\columnwidth}
        \centering
        \includegraphics[width=0.8\columnwidth]{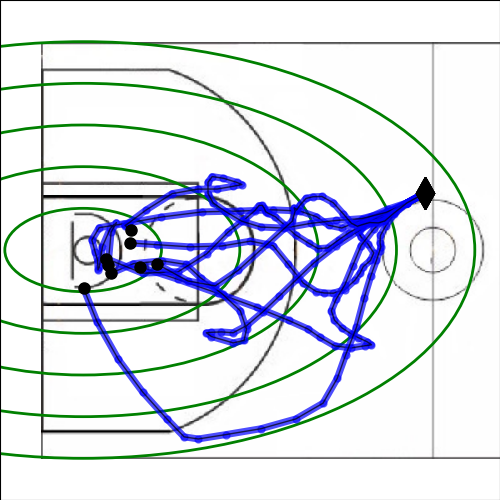}
        \caption{Label class 0 (closest)}
    \end{subfigure}
    \begin{subfigure}[b]{0.32\columnwidth}
        \centering
        \includegraphics[width=0.8\columnwidth]{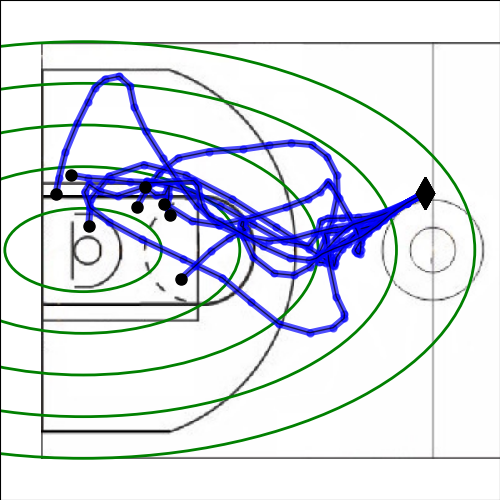}
        \caption{Label class 1}
    \end{subfigure}
    \begin{subfigure}[b]{0.32\columnwidth}
        \centering
        \includegraphics[width=0.8\columnwidth]{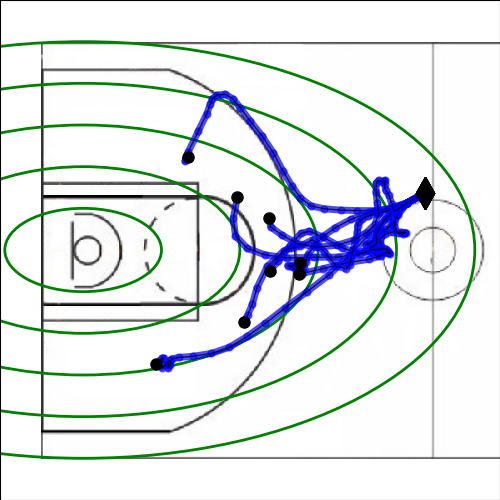}
        \caption{Label class 2}
    \end{subfigure}
    \begin{subfigure}[b]{0.32\columnwidth}
        \centering
        \includegraphics[width=0.8\columnwidth]{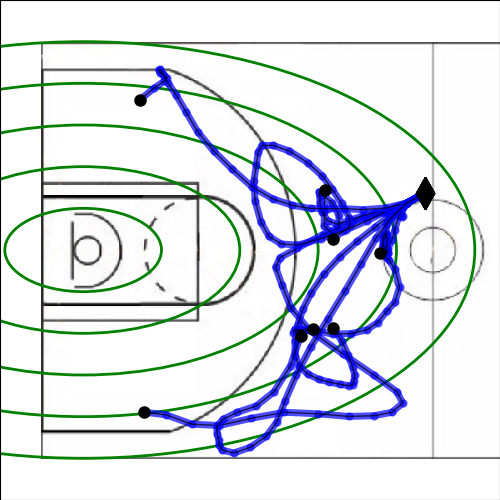}
        \caption{Label class 3}
    \end{subfigure}
    \begin{subfigure}[b]{0.32\columnwidth}
        \centering
        \includegraphics[width=0.8\columnwidth]{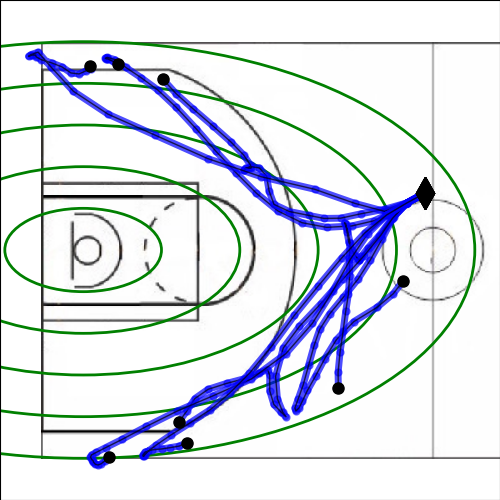}
        \caption{Label class 4}
    \end{subfigure}
    \begin{subfigure}[b]{0.32\columnwidth}
        \centering
        \includegraphics[width=0.8\columnwidth]{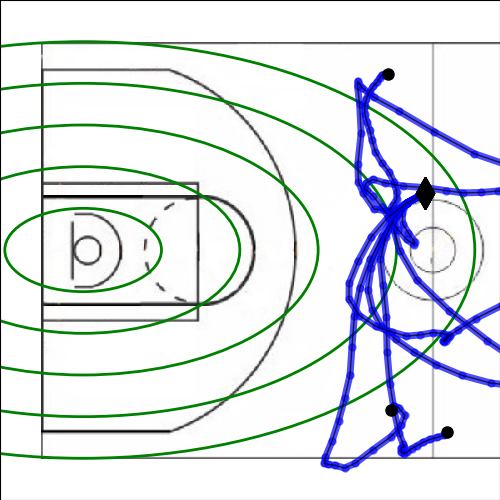}
        \caption{Label class 5 (farthest)}
    \end{subfigure}
\caption{Rollouts from our policy calibrated to \texttt{DESTINATION(net)} with 6 classes. The 5 green boundaries divide the court into 6 regions, each corresponding to a label class. The label indicates the target region of a trajectory's final position ($\bullet$). This policy achieves a \metric{} of 0.93, as indicated in Table \ref{tab:bball_exp2}. Note that the initial position ($\blacklozenge$) is the same as in Figures \ref{fig:bball_lf_dest} and \ref{fig:bball_2lf} for comparison, but in general we sample an initial position from the prior $p(\bfy)$ to compute \metric{}. }
\label{fig:bball_dest6}
\end{figure}

\end{document}